\documentclass[10pt,letterpaper,compsoc,conference]{iiswc26}

\usepackage{cite}
\usepackage{amsmath,amssymb,amsfonts}
\usepackage{algorithmic}
\usepackage{graphicx}
\usepackage[dvipsnames,table]{xcolor}
\usepackage[final]{microtype}
\usepackage[italic]{mathastext}
\usepackage{libertine}
\usepackage[T1]{fontenc}
\usepackage{textcomp}
\usepackage[varqu,varl]{zi4}
\usepackage[all]{nowidow}
\usepackage[keeplastbox]{flushend}
\usepackage{fancyhdr}
\usepackage{enumitem}

\usepackage{algorithm}
\usepackage{multirow}
\usepackage{booktabs}
\usepackage{array}
\usepackage{fontawesome5}
\usepackage{hwemoji}
\usepackage{afterpage}
\usepackage{placeins}
\usepackage{caption}
\usepackage{float}
\usepackage[hidelinks]{hyperref}
\usepackage{url}
\usepackage[most]{tcolorbox} 
\definecolor{improveGreen}{RGB}{229,243,228}
\definecolor{captionGreen}{RGB}{198,233,180}
\definecolor{sectionBg}{RGB}{240,240,240}
\definecolor{rowMA}{RGB}{253,232,229}
\definecolor{rowMB}{RGB}{253,238,218}
\definecolor{rowMC}{RGB}{253,245,210}
\definecolor{rowMD}{RGB}{231,236,245}
\definecolor{medalGold}{RGB}{218,165,32}
\definecolor{medalSilver}{RGB}{160,160,160}
\definecolor{medalBronze}{RGB}{176,98,53}

\makeatletter
\newcommand{\sparkle}{\raisebox{-0.15ex}{\scalebox{1.5}{\hwemoji@xsxr}}}
\makeatother

\definecolor{promptHeader}{RGB}{26,35,90}
\definecolor{promptBody}{RGB}{237,241,252}
\newtcolorbox{promptbox}[1]{
  colback=promptBody,
  colframe=promptHeader,
  coltitle=white,
  fonttitle=\small\rmfamily,
  title=#1,
  boxrule=0.8pt,
  arc=2pt,
  toptitle=1.5mm,
  bottomtitle=1.5mm,
  left=4pt, right=4pt, top=4pt, bottom=4pt
}

\definecolor{takeawayBg}{HTML}{E6EEF8}
\definecolor{takeawayBorder}{HTML}{3A5E9E}
\newcounter{takeawayCounter}
\newcommand{\takeaway}[1]{%
  \par\vspace{0.5em}\noindent
  \refstepcounter{takeawayCounter}%
  {\setlength{\fboxrule}{1.2pt}\setlength{\fboxsep}{8pt}%
   \fcolorbox{takeawayBorder}{takeawayBg}{%
     \begin{minipage}{0.93\columnwidth}\small
       \textbf{Takeaway \#\thetakeawayCounter:}~#1
     \end{minipage}%
   }}\par\vspace{0.5em}
}

\newcommand{\iiswcsubmissionnumber}{349}
\fancypagestyle{firstpage}{
  \fancyhf{}
  
  \fancyhead[C]{\vspace{10pt}\normalsize{IISWC 2026 Submission
      \textbf{\#\iiswcsubmissionnumber} -- Confidential Draft -- Do
      NOT Distribute!!} \\\vspace{-25pt}}
  \fancyfoot[C]{\thepage}
}

\begin{document}

\author{
  \IEEEauthorblockN{Musa Cim, Burak Topcu, Chita Das, Mahmut Kandemir}
  \IEEEauthorblockA{The Pennsylvania State University\\
  \{mtc5693, bvt5283, cxd12, mtk2\}@psu.edu}
}

\title{Parallel Context Compaction for Long-Horizon LLM Agent Serving}

\maketitle
\pagestyle{plain}

\begin{abstract}
Long-horizon LLM agents accumulate growing conversation histories that eventually exceed the model's context window. Context compaction via LLM-based summarization keeps the conversation bounded, but summarization is inherently lossy and the blocking call stalls agent inference for tens of seconds. Moreover, the operator has no fine-grained control over summary volume since prompt instructions are largely ignored, and as context grows, both the amount of output tokens the model produces and the information it retains fluctuate substantially from run to run, making the agent's retained knowledge unpredictable across runs.
We introduce \textbf{parallel compaction} for long-horizon agentic flows and characterize it against the sequential synchronous baseline across four backbones spanning 8B to 120B parameters, mixing dense and MoE architectures with reasoning and non-reasoning models, on the HotpotQA multi-hop QA and LoCoMo long-context dialogue benchmarks.
Parallel compaction gives the operator fine-grained, predictable control over summary volume and enables more targeted prompt engineering per block. At matched compaction decode volume, it reduces end-to-end wall time and improves compaction throughput over the sequential baseline.
\end{abstract}

\section{Introduction}
Large Language Model (LLM) agents are increasingly deployed in long-horizon settings where they process many tasks within a single session, e.g., multi-hop question answering over accumulated evidence, agentic code editing with tool use, and multi-session dialogue. In each setting, the conversation history grows with each interaction, surfacing many practical problems: per-token latency increases with sequence length, the conversation eventually exceeds the model's context window, and reasoning quality degrades well before that window is reached, as the model is forced to attend over an increasingly long, low-signal history. Summarization, the natural remedy, faces its own limits: models struggle to produce summaries of predictable length and content as context grows.

Context compaction periodically summarizes the conversation history into a shorter representation, keeping the context bounded as the agent processes more tasks. Production systems such as Anthropic's Claude Code~\cite{anthropic2025claudecode} and OpenAI's Codex~\cite{openai2025codex}, along with widely used frameworks such as LangChain~\cite{langchain} and LlamaIndex~\cite{llamaindex}, rely on LLM summarization calls to compress context when approaching window limits. This compaction is typically performed synchronously, and because a summary is inherently shorter than the original, information is always lost: the context is reduced by 90--99\% in token count, yet the model decides what to keep and what to drop with no guarantee of consistency across runs. In interactive settings such as CLI coding agents, the information loss is particularly damaging: summarization is so aggressive that the most recent context, which contains the insights the agent has just accumulated, is heavily compressed alongside the rest, forcing the agent to rediscover and re-establish that context in subsequent turns at the cost of extra tokens and time.

Current frontier models support context windows from 100k to 1M tokens (roughly 150–2,500 PDF pages of dense text). Yet feeding this much material into a single inference call yields diminishing returns on most use cases: well-known limitations such as \emph{lost in the middle}~\cite{liu2024lost} and \emph{context rot}~\cite{hong2025contextrot} show that LLMs struggle to attend over long, low-signal histories, and the same limitations apply during summarization; for example, a model asked to compress a 96k-token context attends poorly over the full history, producing short and inconsistent summaries. The summary volume, information retention, and latency cost of context compaction under long-horizon multi-turn agentic flows have not been systematically characterized, neither for the naive sequential baseline nor for parallel approaches. Our contributions are:

\begin{enumerate}%[topsep=1pt, leftmargin=2em]
\item \textbf{Characterization of compaction behavior.} We show empirically that summarization output volume is essentially input- and prompt-invariant across four backbones spanning 8B to 120B parameters. As input grows from 2k to 96k tokens, output grows by only $\sim$3$\times$, and switching the instruction from \emph{concise} to \emph{very detailed} barely moves the output. Moreover, both summary length and content become increasingly unstable at longer contexts, making information loss unpredictable for the agent.

\item \textbf{Parallel compaction design and characterization.} We introduce parallel compaction, a block-based design that gives the operator direct, fine-grained control over summary volume through block count, and characterize it against the sequential baseline on HotpotQA and LoCoMo across four backbones, reporting end-to-end wall time, compaction throughput, and summary volume across a block-size sweep from 16k down to 2k tokens.
\end{enumerate}

%%%%%
\section{Background and Motivation}
\subsection{Long-Horizon LLM Agents}
We consider an LLM agent that processes a sequence of $T$ tasks $(q_1, q_2, \ldots, q_T)$ within a single conversation session. At step $t$, the agent receives task $q_t$ appended to the accumulated conversation history $\mathcal{H}_t = [s, u_1, a_1, \ldots, u_{t-1}, a_{t-1}]$, where $s$ is the system prompt, $u_i$ is the $i$th user message including supporting context, and $a_i$ is the corresponding model response. As more tasks are processed, the growing context length raises per-token decode cost.

\subsection{Context Compaction}
When the context length exceeds a threshold $\tau$, compaction summarizes the conversation history into a shorter representation:
\begin{equation}
\hat{\mathcal{H}}_t = \text{LLM}_{\text{compact}}(\mathcal{H}_t \oplus p_c)
\label{eq:compaction}
\end{equation}
where $p_c$ is a compaction instruction prompt. The agent blocks until the summary is generated, after which the compacted summary replaces the conversation history. The choice of threshold $\tau$ creates a fundamental tradeoff: lower thresholds keep context small but incur more blocking overhead and risk information loss, while higher thresholds preserve more context but allow latency to grow.

\subsection{Motivating Measurements}
Before introducing parallel compaction, we present motivating measurements that characterize the fundamental limitations of sequential compaction --- why prompt engineering and input length alone cannot control summary volume, and how much wall-time overhead synchronous compaction imposes, motivating the need for a different approach.

\subsubsection{Summary Output is Input-Invariant}
\label{sec:scaling}
We measured whether input length can control summary volume. Across four backbones on HotpotQA and LoCoMo, input length grows by 48$\times$ (2k$\to$96k) while average output grows by only 3$\times$ (Table~\ref{tab:scaling-numbers}).

\begin{table}[H]
\centering
%\captionsetup{font=footnotesize}
\caption{Avg.\ summarization output vs.\ input length. Output stays nearly constant as input grows.}
\label{tab:scaling-numbers}
\small
\setlength{\tabcolsep}{4pt}
\begin{tabular}{r r r}
\toprule
\textbf{Input tokens} & \textbf{Avg output tokens} & \textbf{Output / Input} \\
\midrule
2{,}048   &   981   & 47.9\,\% \\
4{,}096   & 1{,}077 & 26.3\,\% \\
8{,}192   & 1{,}297 & 15.8\,\% \\
16{,}384  & 2{,}087 & 12.7\,\% \\
32{,}768  & 1{,}391 & ~4.2\,\% \\
65{,}536  & 2{,}120 & ~3.2\,\% \\
98{,}304  & 3{,}015 & ~3.1\,\% \\
\bottomrule
\end{tabular}
\end{table}

\takeaway{As input length grows 48$\times$ from 2k to 96k tokens, summarization output grows by only $\sim$3$\times$: the model self-bounds its output regardless of how much context it receives.}

\subsubsection{Compaction Dominates Agent Wall Time} 
To quantify the cost of synchronous compaction, we run the agentic flow end-to-end under different context thresholds and measure how much of the total wall time is spent in compaction calls. Table~\ref{tab:compaction-fraction} shows that synchronous compaction dominates wall time at low thresholds, consuming up to 62\% of total execution on HotpotQA, and remains significant even at higher thresholds. The overhead compounds with frequency: at $\tau=16\,$k, compaction fires 15 times per run, each being a full stall, whereas at $\tau=96\,$k, only 2 compactions occur, but each processes a much longer context. Reducing per-call latency is therefore the primary lever for improving end-to-end throughput. 

\begin{table}[H]
\centering
%\captionsetup{font=footnotesize}
\caption{Compaction fraction of E2E wall time vs.\ threshold.}
\label{tab:compaction-fraction}
\small
\setlength{\tabcolsep}{4pt}
\begin{tabular}{l r r r}
\toprule
\textbf{Model} & \textbf{Threshold} & \textbf{\# Compactions} & \textbf{Compact / E2E} \\
\midrule
\multirow{4}{*}{gpt-oss-20B}  & 16\,k & 15 & 51.3\,\% \\
                              & 32\,k &  7 & 28.2\,\% \\
                              & 64\,k &  3 & 14.5\,\% \\
                              & 96\,k &  2 &  8.6\,\% \\
\midrule
\multirow{4}{*}{Llama-3.1-8B} & 16\,k & 15 & 62.4\,\% \\
                              & 32\,k &  7 & 53.6\,\% \\
                              & 64\,k &  3 & 24.4\,\% \\
                              & 96\,k &  2 & 14.5\,\% \\
\bottomrule
\end{tabular}
\end{table}

\takeaway{Synchronous compaction can become a bottleneck in long-horizon agentic flows, consuming a substantial fraction of end-to-end wall time.}

%%%%%

\section{Experimental Methodology}
\label{sec:methodology}

\subsection{Benchmarks}

We evaluate two agentic-flow benchmarks. \textbf{HotpotQA}~\cite{yang2018hotpotqa} is a multi-hop QA dataset with Wikipedia-based question-answer pairs. \textbf{LoCoMo}~\cite{maharana2024evaluating} provides long multi-session conversations with multiple QAs each, using the multiple-choice question format~\cite{percena_locomo_mc10}. We feed documents one at a time and ask each question immediately after it is read, so the conversation accumulates as a long-horizon agent's history would.

For each question, the backbone LLM generates a free-form answer graded by Qwen3-30B as an independent LLM-as-judge using a rubric template, run deterministically and blind to avoid self-preference bias. LLM evaluators tend to favor their own outputs~\cite{panickssery2024llm, xu2024pride}, so using a judge from a different model family than our backbones (GPT and Llama) ensures a fairer evaluation.

\begin{table}[H]
\centering
\caption{Benchmark statistics.}
\label{tab:benchmarks}
\small
\setlength{\tabcolsep}{4pt}
\begin{tabular}{l c c}
\toprule
\textbf{Benchmark} & \textbf{Questions/doc} & \textbf{Doc length} \\
\midrule
HotpotQA & 1 & $\sim$1.2k tokens \\
LoCoMo   & 6 & $\sim$0.7k tokens \\
\bottomrule
\end{tabular}
\end{table}

\begin{figure*}[!t]
\centering
\includegraphics[width=0.99\textwidth]{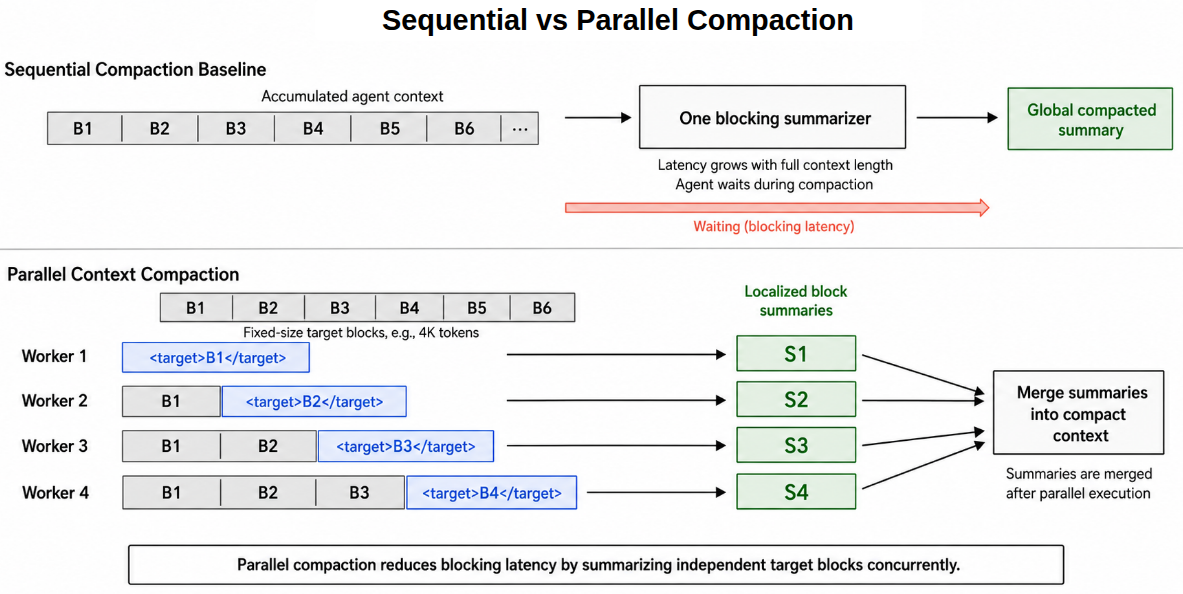}
\caption{Sequential versus parallel compaction overview.} \vspace{-1mm}
\label{fig:architecture}
\end{figure*}

%%%%%

\section{Parallel Compaction}
\label{sec:methods} 
When the conversation length exceeds a compaction threshold $\tau$, parallel compaction proceeds through three phases (Figure~\ref{fig:architecture}):

\begin{enumerate}
\item \textbf{Snapshot \& partition.} Copy the current conversation $\mathcal{X}$, record its length, and partition it into $N$ contiguous blocks of size $B$ tokens each, where $N = \lceil |\mathcal{X}| / B \rceil$ and $B$ is a fixed configuration knob.

\item \textbf{Dispatch.} For each block $k \in \{1,\ldots,N\}$, build a prompt consisting of blocks $1$ through $k$ in order, with block $k$ wrapped in a \texttt{<TARGET\_BLOCK> \ldots </TARGET\_BLOCK>} marker. All $N$ prompts are dispatched concurrently to the vLLM server.

\item \textbf{Merge.} Once all workers complete, the $N$ per-block summaries are concatenated in block order to form the compacted history $\hat{\mathcal{H}}_t$, which replaces the original snapshot.
\end{enumerate}

\noindent \textbf{Prefix-aware target-at-end layout.} Two natural alternatives exist: feed each worker only its block (\emph{independent chunks}), or feed every worker the entire snapshot with a worker-specific marker (\emph{shared full prefix}). Independent chunks break the prefix cache and lose cross-chunk context; shared full prefix breaks it too, since each worker's marker lands at a different offset. The prefix-aware target-at-end layout is the middle ground: the target block sits at the end of the visible prompt where attention is strongest, each worker's prefix strictly extends the previous worker's, preserving the cache, and cross-block context is maintained because each worker sees the full conversation history up to its block. This is a natural fit for agentic flows, where the conversation is built chronologically from left to right, so the prefix always reflects the causal order in which context was accumulated.

\subsection{Models}
We evaluate four backbones to cover the regimes most relevant for production deployments: \textbf{Llama-3.1-8B} (small dense), \textbf{gpt-oss-20B} (medium MoE reasoning), \textbf{Llama-3.3-70B} (large dense), and \textbf{gpt-oss-120B} (large MoE reasoning). The four-backbone span captures small/large, dense/MoE, and reasoning/non-reasoning architectures. Table~\ref{tab:models} lists each model's type, reasoning capability, context window, and GPU configuration. This selection also spans the practical cost spectrum, from a single-GPU deployment accessible to most practitioners to a large MoE model representative of frontier serving costs.

\begin{table}[H]
\centering
\caption{Models used in our experiments.}
\label{tab:models}
\footnotesize
\setlength{\tabcolsep}{3pt}
\begin{tabular}{l l c c l}
\toprule
\textbf{Model} & \textbf{Type} & \textbf{Reasoning} & \textbf{Ctx} & \textbf{Hardware} \\
\midrule
Llama-3.1-8B   & Dense & No  & 128k & 1$\times$ H100 \\
Llama-3.3-70B  & Dense & No  & 128k & 2$\times$ H100 \\
gpt-oss-20B    & MoE   & Yes & 128k & 1$\times$ H100 \\
gpt-oss-120B   & MoE   & Yes & 128k & 1$\times$ H100 \\
\bottomrule
\end{tabular}
\end{table}

\subsection{Hardware and Serving Infrastructure} 
\label{sec:seq-char}
All experiments use vLLM with prefix caching and chunked prefill enabled. Llama-3.1-8B and gpt-oss-20B serve at TP=1 on a single H100; gpt-oss-120B serves at TP=1 with mxfp4 weights on a single H100; Llama-3.3-70B requires TP=2 across an H100 NVL pair. Each measurement runs on a dedicated GPU (or NVLink pair) with no other workloads sharing the device.

\begin{figure*}[!t]
\centering
\begin{minipage}[t]{0.49\textwidth}
  \centering
  \includegraphics[width=1\textwidth]{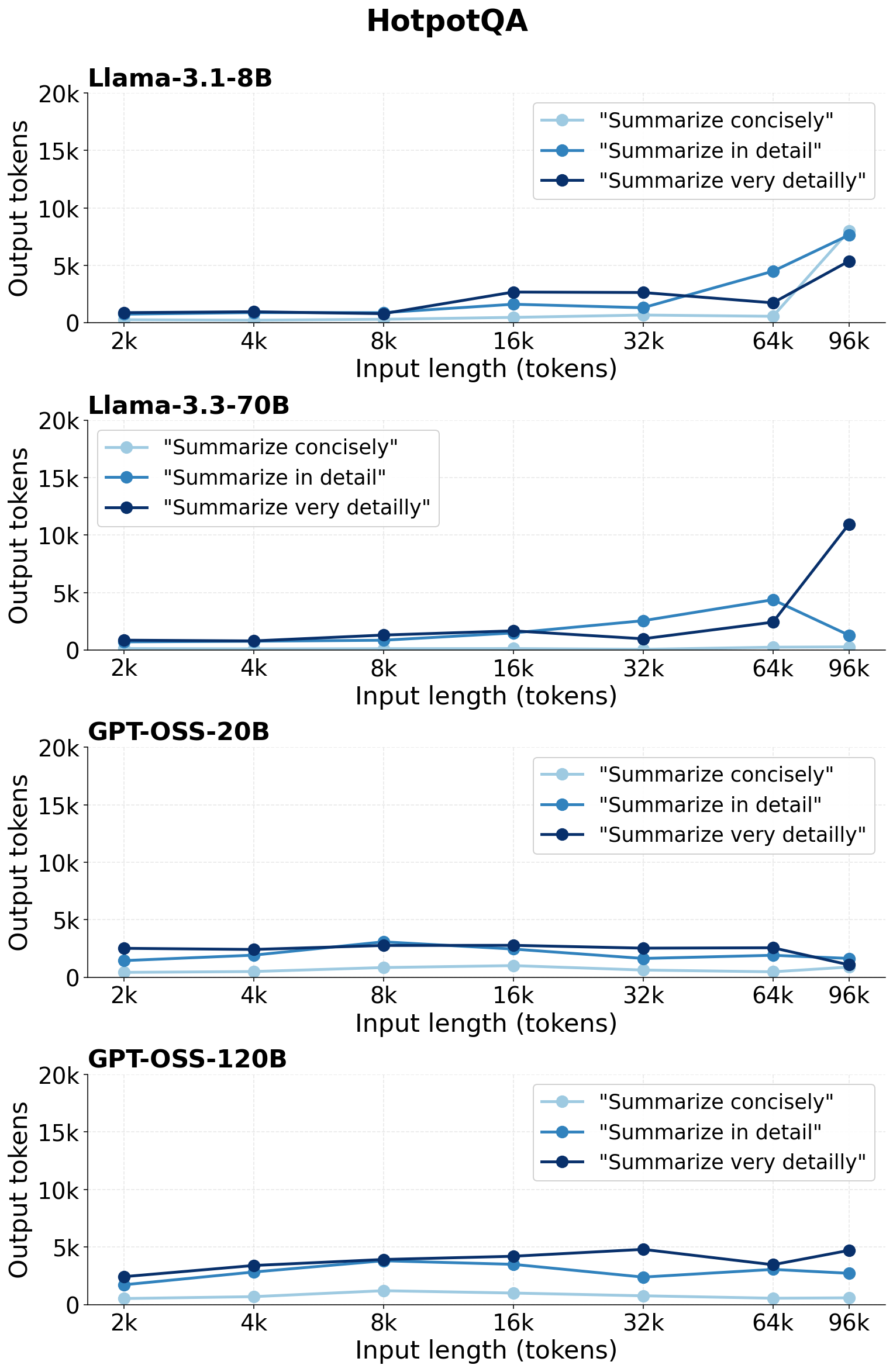}\\
  \small (a) HotpotQA
\end{minipage}\hfill
\begin{minipage}[t]{0.49\textwidth}
  \centering
  \includegraphics[width=1\textwidth]{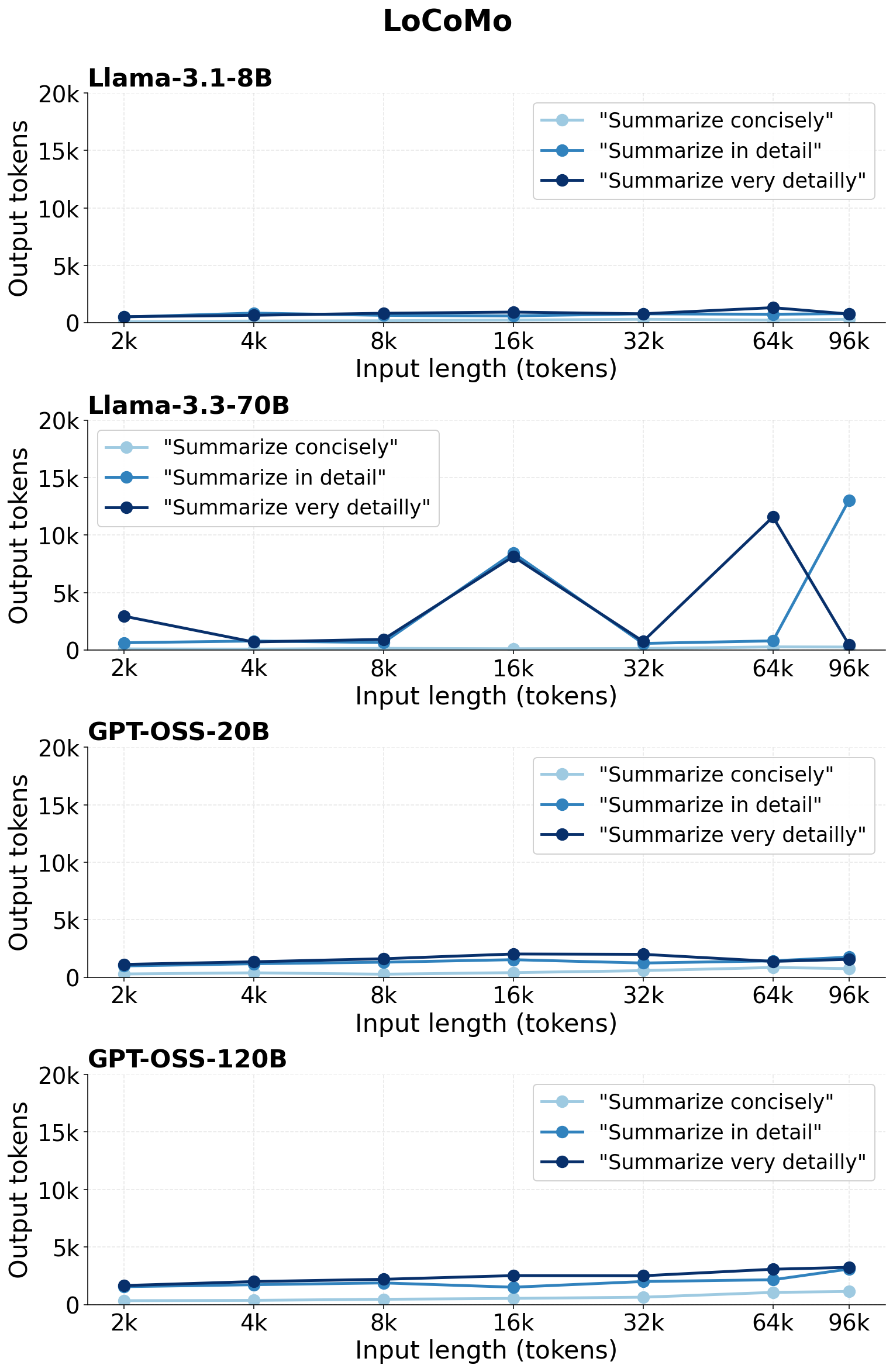}\\
  \small (b) LoCoMo
\end{minipage}
%\captionsetup{font=footnotesize}
\vspace{-1.5mm}
\caption{Output tokens vs.\ input length for three prompt variants and four backbones under Sequential compaction.}
\vspace{-3.5mm}
\label{fig:scaling}
\end{figure*}

\section{Characterizing Sequential Compaction}
Sequential compaction issues a single blocking summarization call over the entire conversation snapshot whenever the context exceeds the threshold $\tau$. We characterize this baseline along two axes: output volume scaling, measuring how much the model produces as the input grows and as the prompt changes, and run-to-run stability, measuring how consistent that output is across repeated calls with the same input.

\subsection{Prompt and Input Invariance}
Figure~\ref{fig:scaling} plots summarization output tokens against input length for all four backbones on HotpotQA (a) and LoCoMo (b), under three prompt variants. Two patterns dominate. First, all four backbones produce output tokens in a narrow range across the full 2\,k to 96\,k input sweep, which is far flatter than the 48$\times$ growth in input length would suggest. Second, switching the prompt from \emph{concise} to \emph{very detailed} barely moves the curves.

This behavior is consistent across dense and MoE architectures, as well as across reasoning and non-reasoning models on both benchmarks. Llama-3.3-70B tends to produce larger outputs than the other backbones at the same input length, but its behavior is non-deterministic across runs, making its output volume harder to predict. The operator has no effective prompt-side knob to grow summary volume as the conversation grows longer. The practical consequence is that sequential compaction progressively discards more and more of the conversation as context grows: at 2\,k input, the model retains roughly 48\% of the tokens in the summary, but at 96\,k, that figure drops to $\sim$3\%. 

\takeaway{Changing the prompt has no significant impact on output length: models largely ignore length instructions and self-bound their output regardless of whether the prompt asks for a concise or very detailed summary. The operator cannot use prompt engineering to control compaction volume.}
 
\subsection{Run-to-Run Instability}
Beyond input invariance, both output length and content vary substantially from run to run. This is a fundamental but under-characterized property of LLM-based compaction: given an identical conversation snapshot and the same prompt, the model produces a different summary on every call. The output length may shift by hundreds of tokens, and the semantic content of the summary changes accordingly. This matters because the compacted history directly conditions all subsequent reasoning steps---an agent that receives a shorter or differently-framed summary on one run will follow a different reasoning path than on another, even though the underlying task is identical. Instability worsens with context length: longer inputs give the model more room to vary which facts it selects, how it phrases them, and how much it elaborates, amplifying the token-count and semantic spread across runs. To systematically characterize this behavior, we sweep input length from 4\,k to 96\,k tokens and, at each point, run compaction ten times under identical prompt and input conditions with temperature-driven sampling, measuring how much the output length varies and how similar the generated summaries are to one another.

\noindent \textbf{Output length variability: coefficient of variation.} To quantify how much the output token count shifts across runs, we use the coefficient of variation (CV), defined as the ratio of the standard deviation to the mean of the output length across the ten repeated compaction calls. A CV of 0\% indicates perfectly consistent output length; a CV of 100\% means the standard deviation equals the mean, reflecting a very wide spread relative to the average output size. We prefer CV over raw standard deviation because it is scale-independent --- a CV of 30\% means the same relative spread whether the model produces 100 or 5{,}000 tokens on average, making it comparable across models and input lengths. In the compaction context, high CV means that the operator cannot predict how many tokens will be consumed by the summary on any given run, which makes memory budgeting and context threshold planning unreliable.

\noindent \textbf{Summary content variability: cosine similarity.} Token count alone does not capture whether two summaries actually say the same thing --- a model could produce outputs of identical length while selecting entirely different facts. To measure semantic consistency, we embed each of the ten summaries using a fixed sentence encoder and compute the average pairwise cosine similarity across all pairs. A cosine similarity of 1.0 means every run produces semantically identical text; values near 0 indicate that runs share almost no semantic content. Low cosine similarity is particularly damaging in agentic flows: if the compacted summary omits a key fact in one run but retains it in another, downstream QA accuracy will vary between runs not because the model reasoned differently, but because it received a different context—making performance non-reproducible even on a fixed benchmark.

% Requires in preamble:
%   \usepackage[dvipsnames,table]{xcolor}
%   \usepackage{multirow}
%   \usepackage{booktabs}
% Colors expected from main.tex: improveGreen, captionGreen, sectionBg, rowMA..rowMD.

% ─────────────────────────────────────────────────────────────────────────
% TABLE 1: Stability across input sizes (single-column, 4k → 96k sweep, Δ vs 4k)
% ─────────────────────────────────────────────────────────────────────────
\begin{table}[h]
    \centering
    %\captionsetup{font=footnotesize}
    \caption{Stability across input sizes (4--96\,k).}
    \label{tab:stability-vs-input-size}
    \scriptsize
    \setlength{\tabcolsep}{3pt}
    \renewcommand{\arraystretch}{1.15}
    \resizebox{\columnwidth}{!}{%
    \begin{tabular}{ll|rc|rc}
    \toprule
    \multirow{2}{*}{\textbf{Model}} & \multirow{2}{*}{\textbf{Input}}
     & \multicolumn{2}{c|}{\textbf{Coefficient of Variation (\%)} $\downarrow$}
     & \multicolumn{2}{c}{\textbf{Cosine Similarity} $\uparrow$} \\
    \cmidrule(lr){3-4}\cmidrule(lr){5-6}
     & & \cellcolor{white}Value & \cellcolor{captionGreen}\textbf{$\Delta$}
       & \cellcolor{white}Value & \cellcolor{captionGreen}\textbf{$\Delta$} \\
    \midrule

    % gpt-oss-20B (baseline 4k: CV 24.5%, cos 0.777)
    \rowcolor{rowMA}
     & \cellcolor{white}4\,k  & \cellcolor{white}24.5\%  & \cellcolor{improveGreen}---
     & \cellcolor{white}0.777 & \cellcolor{improveGreen}--- \\
    \rowcolor{rowMA}
     & \cellcolor{white}8\,k  & \cellcolor{white}21.5\%  & \cellcolor{improveGreen}\textcolor{ForestGreen}{$-3.0$}
     & \cellcolor{white}0.741 & \cellcolor{improveGreen}\textcolor{BrickRed}{$-0.036$} \\
    \rowcolor{rowMA}
     & \cellcolor{white}16\,k & \cellcolor{white}38.7\%  & \cellcolor{improveGreen}\textcolor{BrickRed}{$+14.2$}
     & \cellcolor{white}0.726 & \cellcolor{improveGreen}\textcolor{BrickRed}{$-0.051$} \\
    \rowcolor{rowMA}
     & \cellcolor{white}32\,k & \cellcolor{white}27.3\%  & \cellcolor{improveGreen}\textcolor{BrickRed}{$+2.8$}
     & \cellcolor{white}0.654 & \cellcolor{improveGreen}\textcolor{BrickRed}{$-0.123$} \\
    \rowcolor{rowMA}
     & \cellcolor{white}64\,k & \cellcolor{white}42.0\%  & \cellcolor{improveGreen}\textcolor{BrickRed}{$+17.5$}
     & \cellcolor{white}0.713 & \cellcolor{improveGreen}\textcolor{BrickRed}{$-0.064$} \\
    \rowcolor{rowMA}
    \multirow{-6}{*}{\textbf{gpt-oss-20B}}
     & \cellcolor{white}96\,k & \cellcolor{white}42.9\%  & \cellcolor{improveGreen}\textcolor{BrickRed}{$+18.4$}
     & \cellcolor{white}0.624 & \cellcolor{improveGreen}\textcolor{BrickRed}{$-0.153$} \\
    \midrule

    % gpt-oss-120B (baseline 4k: CV 19.8%, cos 0.825)
    \rowcolor{rowMB}
     & \cellcolor{white}4\,k  & \cellcolor{white}19.8\%  & \cellcolor{improveGreen}---
     & \cellcolor{white}0.825 & \cellcolor{improveGreen}--- \\
    \rowcolor{rowMB}
     & \cellcolor{white}8\,k  & \cellcolor{white}36.0\%  & \cellcolor{improveGreen}\textcolor{BrickRed}{$+16.2$}
     & \cellcolor{white}0.857 & \cellcolor{improveGreen}\textcolor{ForestGreen}{$+0.032$} \\
    \rowcolor{rowMB}
     & \cellcolor{white}16\,k & \cellcolor{white}16.2\%  & \cellcolor{improveGreen}\textcolor{ForestGreen}{$-3.6$}
     & \cellcolor{white}0.783 & \cellcolor{improveGreen}\textcolor{BrickRed}{$-0.042$} \\
    \rowcolor{rowMB}
     & \cellcolor{white}32\,k & \cellcolor{white}70.8\%  & \cellcolor{improveGreen}\textcolor{BrickRed}{$+51.0$}
     & \cellcolor{white}0.751 & \cellcolor{improveGreen}\textcolor{BrickRed}{$-0.074$} \\
    \rowcolor{rowMB}
     & \cellcolor{white}64\,k & \cellcolor{white}62.0\%  & \cellcolor{improveGreen}\textcolor{BrickRed}{$+42.2$}
     & \cellcolor{white}0.760 & \cellcolor{improveGreen}\textcolor{BrickRed}{$-0.065$} \\
    \rowcolor{rowMB}
    \multirow{-6}{*}{\textbf{gpt-oss-120B}}
     & \cellcolor{white}96\,k & \cellcolor{white}68.6\%  & \cellcolor{improveGreen}\textcolor{BrickRed}{$+48.8$}
     & \cellcolor{white}0.619 & \cellcolor{improveGreen}\textcolor{BrickRed}{$-0.206$} \\
    \midrule

    % Llama-3.1-8B (baseline 4k: CV 34.1%, cos 0.652)
    \rowcolor{rowMC}
     & \cellcolor{white}4\,k  & \cellcolor{white}34.1\%  & \cellcolor{improveGreen}---
     & \cellcolor{white}0.652 & \cellcolor{improveGreen}--- \\
    \rowcolor{rowMC}
     & \cellcolor{white}8\,k  & \cellcolor{white}33.0\%  & \cellcolor{improveGreen}\textcolor{ForestGreen}{$-1.1$}
     & \cellcolor{white}0.608 & \cellcolor{improveGreen}\textcolor{BrickRed}{$-0.044$} \\
    \rowcolor{rowMC}
     & \cellcolor{white}16\,k & \cellcolor{white}26.4\%  & \cellcolor{improveGreen}\textcolor{ForestGreen}{$-7.7$}
     & \cellcolor{white}0.576 & \cellcolor{improveGreen}\textcolor{BrickRed}{$-0.076$} \\
    \rowcolor{rowMC}
     & \cellcolor{white}32\,k & \cellcolor{white}48.5\%  & \cellcolor{improveGreen}\textcolor{BrickRed}{$+14.4$}
     & \cellcolor{white}0.665 & \cellcolor{improveGreen}\textcolor{ForestGreen}{$+0.013$} \\
    \rowcolor{rowMC}
     & \cellcolor{white}64\,k & \cellcolor{white}54.9\%  & \cellcolor{improveGreen}\textcolor{BrickRed}{$+20.8$}
     & \cellcolor{white}0.550 & \cellcolor{improveGreen}\textcolor{BrickRed}{$-0.102$} \\
    \rowcolor{rowMC}
    \multirow{-6}{*}{\textbf{Llama-3.1-8B}}
     & \cellcolor{white}96\,k & \cellcolor{white}84.5\%  & \cellcolor{improveGreen}\textcolor{BrickRed}{$+50.4$}
     & \cellcolor{white}0.472 & \cellcolor{improveGreen}\textcolor{BrickRed}{$-0.180$} \\
    \midrule

    % Llama-3.3-70B (baseline 4k: CV 24.4%, cos 0.729)
    \rowcolor{rowMD}
     & \cellcolor{white}4\,k  & \cellcolor{white}24.4\%  & \cellcolor{improveGreen}---
     & \cellcolor{white}0.729 & \cellcolor{improveGreen}--- \\
    \rowcolor{rowMD}
     & \cellcolor{white}8\,k  & \cellcolor{white}41.3\%  & \cellcolor{improveGreen}\textcolor{BrickRed}{$+16.9$}
     & \cellcolor{white}0.620 & \cellcolor{improveGreen}\textcolor{BrickRed}{$-0.109$} \\
    \rowcolor{rowMD}
     & \cellcolor{white}16\,k & \cellcolor{white}35.4\%  & \cellcolor{improveGreen}\textcolor{BrickRed}{$+11.0$}
     & \cellcolor{white}0.590 & \cellcolor{improveGreen}\textcolor{BrickRed}{$-0.139$} \\
    \rowcolor{rowMD}
     & \cellcolor{white}32\,k & \cellcolor{white}47.7\%  & \cellcolor{improveGreen}\textcolor{BrickRed}{$+23.3$}
     & \cellcolor{white}0.630 & \cellcolor{improveGreen}\textcolor{BrickRed}{$-0.099$} \\
    \rowcolor{rowMD}
     & \cellcolor{white}64\,k & \cellcolor{white}26.4\%  & \cellcolor{improveGreen}\textcolor{BrickRed}{$+2.0$}
     & \cellcolor{white}0.579 & \cellcolor{improveGreen}\textcolor{BrickRed}{$-0.150$} \\
    \rowcolor{rowMD}
    \multirow{-6}{*}{\textbf{Llama-3.3-70B}}
     & \cellcolor{white}96\,k & \cellcolor{white}171.6\% & \cellcolor{improveGreen}\textcolor{BrickRed}{$+147.2$}
     & \cellcolor{white}0.491 & \cellcolor{improveGreen}\textcolor{BrickRed}{$-0.238$} \\
    \bottomrule
    \end{tabular}%
    }
\end{table}

% ─────────────────────────────────────────────────────────────────────────

\noindent \textbf{Block size as a stability driver.} The CV and cosine similarity measurements above serve a concrete selection purpose: identifying which block size yields compaction outputs that are stable enough to be reliable in production. Smaller block sizes engage more workers, each summarizing a shorter, more focused segment, which reduces the model's degrees of freedom and produces more consistent outputs from run to run. Larger blocks force the model to compress more content in a single call, increasing the variance in what it chooses to retain and how it phrases it. Block size is therefore not only a throughput lever but also a stability driver.

Across all models in Table~\ref{tab:stability-vs-input-size}, the 4\,k input point is the most favorable: CV ranges from 19.8\% to 34.1\% and cosine similarity stays between 0.65 and 0.83 across all backbones. Some models show slightly better cosine similarity or lower CV at other input sizes, though these gains are not consistent across backbones. In general, as input grows beyond 4\,k, both metrics tend to degrade with no sustained recovery at larger sizes. This makes 4\,k a natural operating point: short enough to limit drift, yet large enough to capture meaningful context. In the parallel compaction setting, a 4\,k block size means each worker's target segment is 4\,k tokens, wrapped in an XML marker, so each individual summarization call operates at the input length where models are most stable.

\takeaway{Run-to-run instability is intrinsic to LLM-based compaction and compounds as context grows --- both output length and summary content become increasingly unpredictable.} 

\begin{figure}[H]
\centering
\includegraphics[width=0.42\textwidth]{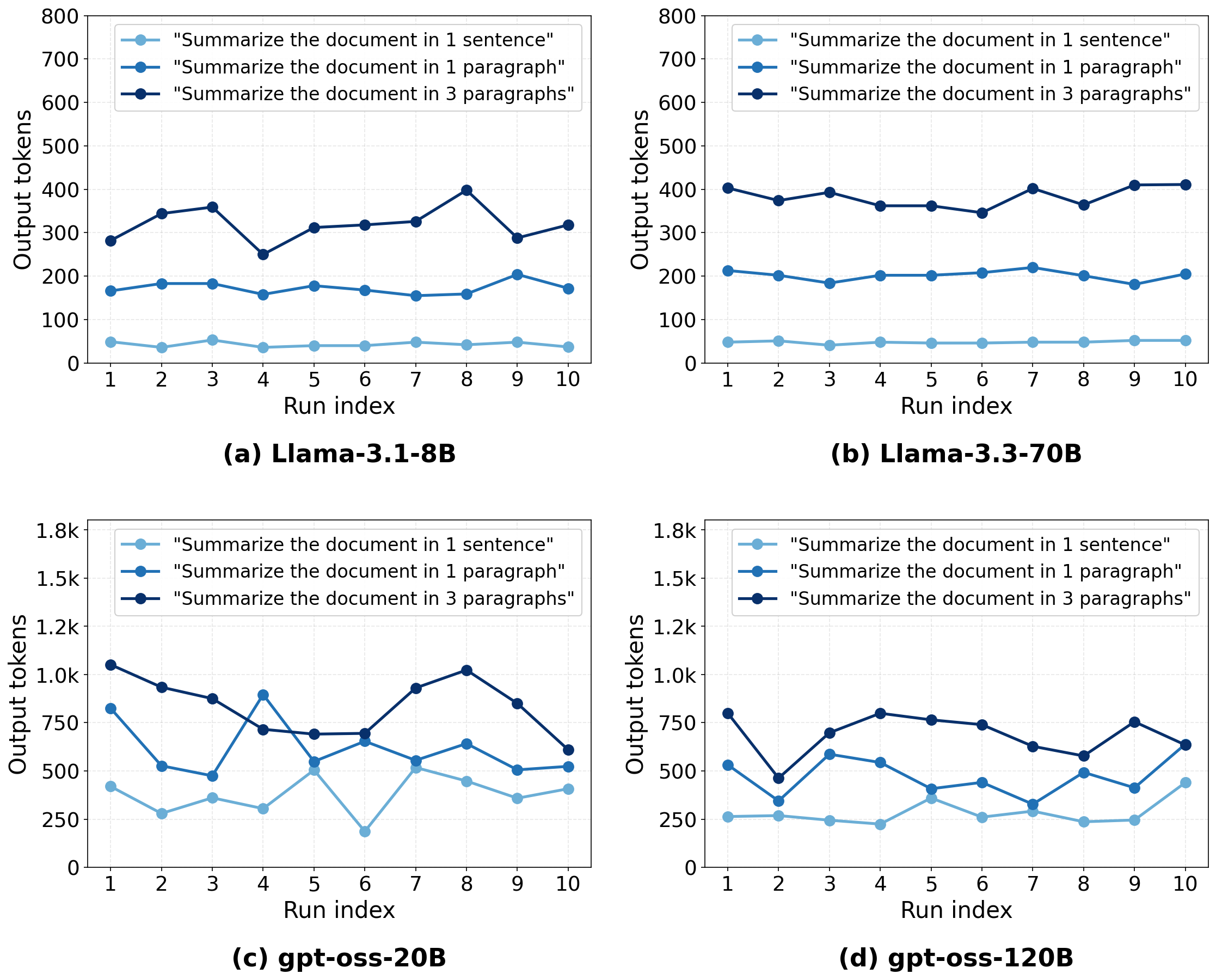}
%\captionsetup{font=footnotesize}
\caption{Output tokens per run across 10 repeated runs at a fixed 4\,k input for three prompt-length variants (light to dark = shorter to longer requested summary length).}
\vspace{-2mm}
\label{fig:prompt-stability-per-run}
\end{figure}

Having established that instability grows with input length, we now fix the input at 4\,k --- the most stable point identified in Table~\ref{tab:stability-vs-input-size} --- and examine how models respond to different prompt-length instructions. We vary the prompt across three levels of requested detail: summarize with one sentence, one paragraph, and three paragraphs. The goal is to assess whether longer prompts can reduce output variability or whether instability persists regardless of the prompt. Figure~\ref{fig:prompt-stability-per-run} shows output token counts across ten repeated runs per model and prompt variant.

Llama instruct models are noticeably more consistent across runs than gpt-oss models, with tighter scatter within each prompt variant. All four backbones scale output tokens from roughly 400 to 600 tokens as the prompt requests more detail, showing that models do respond to prompt-length instructions at this input size. The narrow output range across all models and prompts suggests that models have a strong prior on summary length.

\noindent\textbf{Why LLMs produce short summaries.} A 500-token summary corresponds roughly to half a page to one page of dense text, a length that aligns well with summaries found in standard NLP summarization corpora. Many models are trained or instruction-tuned on document --- summary datasets where reference summaries are often in this range, so they may reproduce this learned length prior at inference time, regardless of how long the source document is. This provides one explanation for why neither input length nor prompt instructions significantly move the output: the model tends to revert to a summary length internalized from training data. As a result, a 500-token summary at 96\,k input retains less than 1\% of the original context.\cite{ravaut2024context}

\takeaway{Summarization output is bounded by the model's internalized prior on summary length: during training, models absorb not just what a summary should contain but also how long it should be, anchoring output length to the distribution of reference summaries in training data regardless of how much input context is provided.}

Table~\ref{tab:stability-vs-prompt} quantifies run-to-run stability across the three prompt-length variants at a fixed 4\,k input. CV and cosine similarity values are largely consistent across variants for each model, with no backbone showing a large or monotonic improvement as the prompt becomes more detailed. Llama-3.3-70B has the lowest CV across all variants, consistent with the tighter scatter in Figure~\ref{fig:prompt-stability-per-run}.

% TABLE 2: Stability across prompt-length variants at fixed 4k input (single-column)
% ─────────────────────────────────────────────────────────────────────────
\begin{table}[H]
    \centering
    \vspace{-2mm}
%    \captionsetup{font=footnotesize}
    \caption{Stability across prompt-length variants at fixed 4\,k input.} 
    \label{tab:stability-vs-prompt}
    \scriptsize
    \setlength{\tabcolsep}{3pt}
    \renewcommand{\arraystretch}{1.15}
    \resizebox{\columnwidth}{!}{%
    \begin{tabular}{ll|r|r}
    \toprule
    \textbf{Model} & \textbf{Prompt}
     & \textbf{CV (\%)} $\downarrow$
     & \textbf{Cosine} $\uparrow$ \\
    \midrule

    \rowcolor{rowMA}
     & \cellcolor{white}Summarize with 1 sentence   & \cellcolor{white}27.2\% & \cellcolor{white}0.704 \\
    \rowcolor{rowMA}
     & \cellcolor{white}Summarize with 1 paragraph  & \cellcolor{white}23.0\% & \cellcolor{white}0.747 \\
    \rowcolor{rowMA}
    \multirow{-3}{*}{\textbf{gpt-oss-20B}}
     & \cellcolor{white}Summarize with 3 paragraphs & \cellcolor{white}18.1\% & \cellcolor{white}0.788 \\
    \midrule

    \rowcolor{rowMB}
     & \cellcolor{white}Summarize with 1 sentence   & \cellcolor{white}23.5\% & \cellcolor{white}0.732 \\
    \rowcolor{rowMB}
     & \cellcolor{white}Summarize with 1 paragraph  & \cellcolor{white}21.8\% & \cellcolor{white}0.819 \\
    \rowcolor{rowMB}
    \multirow{-3}{*}{\textbf{gpt-oss-120B}}
     & \cellcolor{white}Summarize with 3 paragraphs & \cellcolor{white}15.9\% & \cellcolor{white}0.820 \\
    \midrule

    \rowcolor{rowMC}
     & \cellcolor{white}Summarize with 1 sentence   & \cellcolor{white}14.3\% & \cellcolor{white}0.815 \\
    \rowcolor{rowMC}
     & \cellcolor{white}Summarize with 1 paragraph  & \cellcolor{white}8.6\%  & \cellcolor{white}0.719 \\
    \rowcolor{rowMC}
    \multirow{-3}{*}{\textbf{Llama-3.1-8B}}
     & \cellcolor{white}Summarize with 3 paragraphs & \cellcolor{white}13.1\% & \cellcolor{white}0.718 \\
    \midrule

    \rowcolor{rowMD}
     & \cellcolor{white}Summarize with 1 sentence   & \cellcolor{white}6.9\%  & \cellcolor{white}0.922 \\
    \rowcolor{rowMD}
     & \cellcolor{white}Summarize with 1 paragraph  & \cellcolor{white}5.9\%  & \cellcolor{white}0.799 \\
    \rowcolor{rowMD}
    \multirow{-3}{*}{\textbf{Llama-3.3-70B}}
     & \cellcolor{white}Summarize with 3 paragraphs & \cellcolor{white}6.2\%  & \cellcolor{white}0.831 \\
    \bottomrule
    \end{tabular}%
    }\vspace{-2mm}
\end{table}

\noindent \textbf{Long-horizon compaction as long document summarization.} The structure of a long-horizon agentic conversation closely mirrors that of a long document: it is built chronologically, turn by turn, from left to right, with each exchange adding context sequentially. This framing motivates treating compaction as a long document summarization problem, where divide-and-conquer approaches split the input into chunks and summarize each independently. Our parallel compaction design follows this intuition but adds a critical constraint: each block worker receives the full conversation history up to its block rather than just the block itself, preserving cross-block context dependencies and enabling prefix cache reuse across workers --- properties that a naive chunking approach would not preserve. 

\noindent \textbf{Role of the XML marker.} In the prefix-aware layout, each worker receives the full conversation history up to its block, which raises an ambiguity: without explicit delimiters, the model would attempt to summarize the entire prefix rather than its assigned block. The XML marker resolves this by clearly delineating the target region at the end of the prompt, where attention is strongest. This allows the model to attend to the full preceding context for cross-block coherence while focusing its summarization output on the marked segment only.

\noindent \textbf{Threshold and block size interaction.} The compaction threshold $\tau$ and block size $B$ together determine how many workers fire per compaction event: $N = \lceil \tau / B \rceil$. At a fixed threshold of $\tau = 96$\,k tokens, reducing the block size from 16\,k to 4\,k increases the number of parallel workers from 6 to 24. The operator therefore has two independent knobs: $\tau$ controls how often compaction fires, and $B$ controls how many workers run and how much output is produced per event.

\label{sec:par-char}

\begin{figure*}[!t]
\centering
\begin{minipage}[t]{0.46\textwidth}
  \centering
  \includegraphics[width=\linewidth]{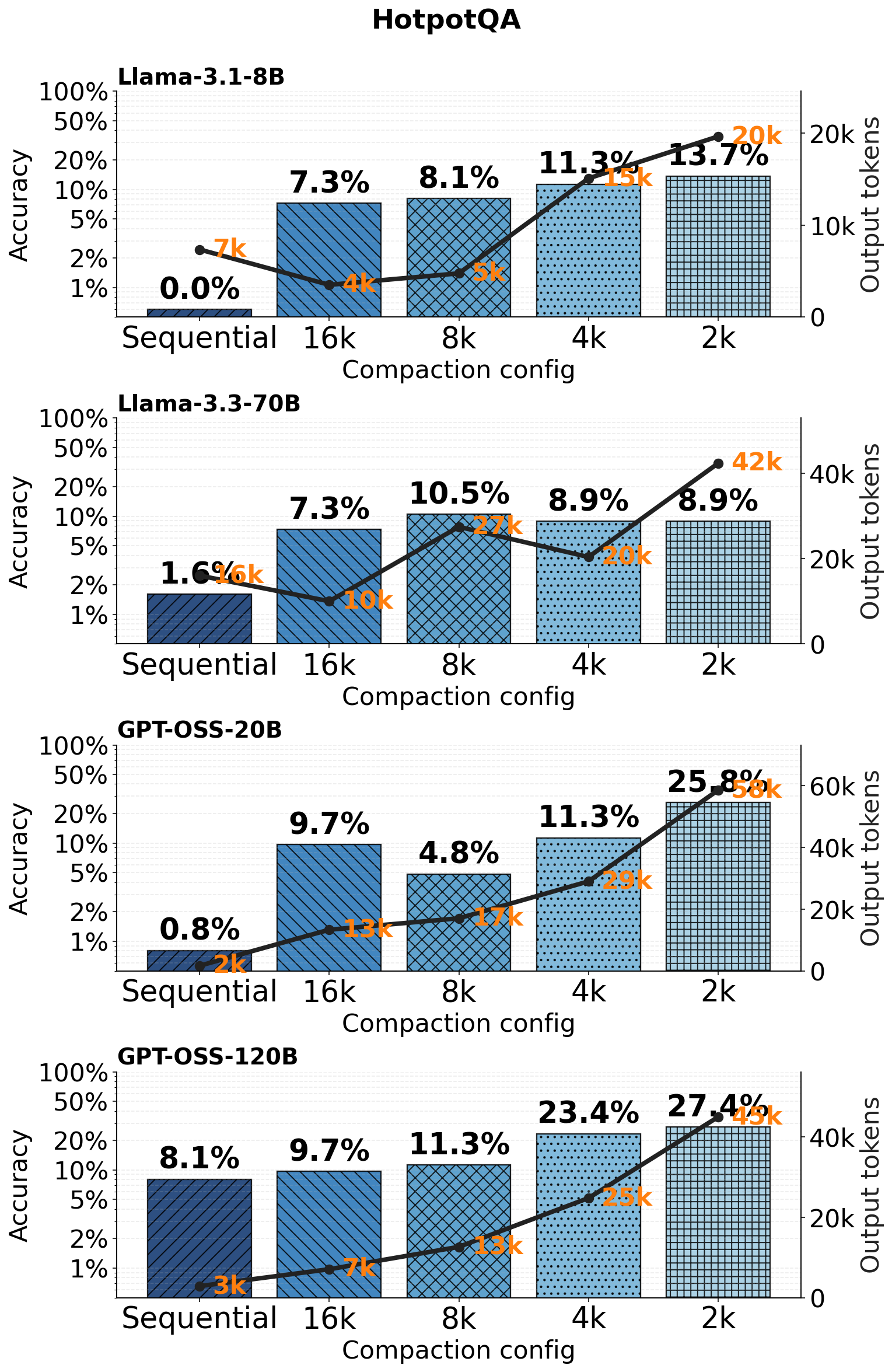}\\
  \small (a) HotpotQA
\end{minipage}\hfill
\begin{minipage}[t]{0.46\textwidth}
  \centering
  \includegraphics[width=\linewidth]{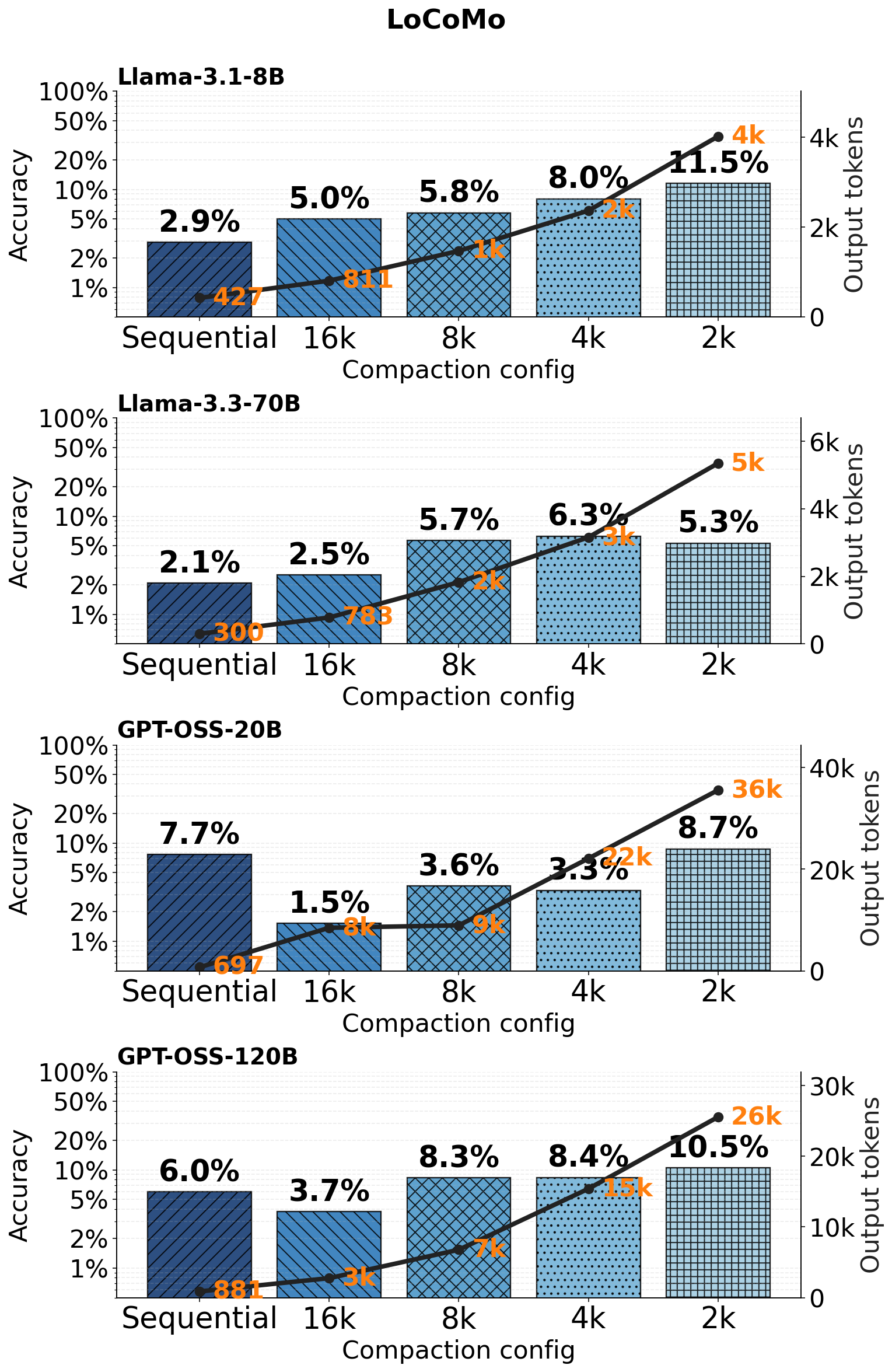}\\
  \small (b) LoCoMo
\end{minipage}
%\captionsetup{font=footnotesize}
\caption{Accuracy per model and configuration on (a) HotpotQA and (b) LoCoMo.}
\vspace{-3mm}
\label{fig:accuracy}
\end{figure*}

\section{Characterizing Parallel Compaction}

Figure~\ref{fig:accuracy} reports accuracy and compaction output token count per configuration on HotpotQA and LoCoMo, with Sequential and each block size on the x-axis, accuracy as histogram bars, and output tokens overlaid. For each configuration, we feed a 100\,k-token context to the model, ask it to summarize, and use Qwen3-30B as a judge to assess whether the summary preserves the information needed to answer the corresponding questions. The 16\,k block size produces the fewest output tokens; as block size decreases, more workers run in parallel, total output tokens increase, and more of the original context is preserved in the summary. Accuracy improves over the Sequential baseline as block size decreases, and this relationship is approximately linear --- as total compaction output grows with the number of workers, accuracy increases monotonically across the block-size sweep, consistently across all configurations. It is worth noting that accuracy does not reach 100\% even at the smallest block size, as performance is also bounded by the model's own reasoning capability on the downstream task, not solely by information preservation in the summary.

\takeaway{Block size directly controls how much context survives compaction, improving accuracy over sequential by preserving more context in the summary.}

Having established that parallel compaction improves accuracy over sequential, we now evaluate its system-level performance in a multi-turn agentic flow. The agent processes a sequence of QA tasks, accumulates context across turns, and triggers compaction at a fixed threshold of $\tau=96$\,k tokens. Table~\ref{tab:speedup-hotpot} reports end-to-end wall time, throughput, compaction decode tokens, QA decode tokens, and average compaction throughput. Compaction decode tokens measure how many tokens the compaction workers generate, while QA decode tokens capture the agent's downstream answer generation. Average compaction throughput accounts for both prefill and decode cost of the compaction.

Table~\ref{tab:speedup-hotpot} shows end-to-end speedups, though measuring a fair throughput comparison is non-trivial: compaction output token counts fluctuate across runs and LLMs cannot reliably be constrained to a specific length, making byte-identical token counts across configurations difficult to achieve. 

% Requires in preamble:
%   \usepackage[dvipsnames,table]{xcolor}
%   \usepackage{multirow}
%   \usepackage{booktabs}
% Colors expected from main.tex: improveGreen, captionGreen, sectionBg, rowMA..rowMD.

\begin{table*}[!tbp]
  \centering
  \caption{\textbf{HotpotQA: Sequential vs Parallel across block sizes.} Fixed threshold $\tau=96\text{k}$. Sequential is the single-call blocking compaction baseline; Parallel uses one worker per block (block size sweep: 16k, 8k, 4k, 2k tokens; smaller block $\Rightarrow$ more workers). The \colorbox{captionGreen}{\textbf{$\Delta$}} column shows the speedup ratio over Sequential. Green $=$ improvement, red $=$ regression.}
  \label{tab:speedup-hotpot}
  \scriptsize
  \setlength{\tabcolsep}{2pt}
  \renewcommand{\arraystretch}{1.15}
  \resizebox{\textwidth}{!}{%
  \begin{tabular}{lll|rc|rc|rc|rc|rc}
  \toprule
  \multirow{2}{*}{\textbf{Model}} & \multirow{2}{*}{\textbf{Variant}} & \multirow{2}{*}{\textbf{Block size}}
   & \multicolumn{2}{c|}{\textbf{E2E Wall (s)} $\downarrow$}
   & \multicolumn{2}{c|}{\textbf{E2E Throughput (tok/s)} $\uparrow$}
   & \multicolumn{2}{c|}{\textbf{Compaction Decode (tok)}}
   & \multicolumn{2}{c|}{\textbf{QA Decode (tok)}}
   & \multicolumn{2}{c}{\textbf{Compaction Throughput (ms/tok)} $\downarrow$} \\
  \cmidrule(lr){4-5}\cmidrule(lr){6-7}\cmidrule(lr){8-9}\cmidrule(lr){10-11}\cmidrule(lr){12-13}
   & & & \cellcolor{white}Value & \cellcolor{captionGreen}\textbf{$\Delta$}
     & \cellcolor{white}Value & \cellcolor{captionGreen}\textbf{$\Delta$}
     & \cellcolor{white}Value & \cellcolor{captionGreen}\textbf{$\Delta$}
     & \cellcolor{white}Value & \cellcolor{captionGreen}\textbf{$\Delta$}
     & \cellcolor{white}Value & \cellcolor{captionGreen}\textbf{$\Delta$} \\
  \midrule
  \rowcolor{sectionBg}\multicolumn{13}{c}{\textit{HotpotQA at$\tau=96$k}} \\
  \midrule

  % ─── gpt-oss-20B ───────────────────────────────────────────────────────
  \rowcolor{rowMA}
   & \cellcolor{white}Sequential & \cellcolor{white}—
   & \cellcolor{white}86.9 & \cellcolor{improveGreen}$1.00\times$
   & \cellcolor{white}145.2 & \cellcolor{improveGreen}$1.00\times$
   & \cellcolor{white}1{,}493 & \cellcolor{improveGreen}$1.00\times$
   & \cellcolor{white}11{,}127 & \cellcolor{improveGreen}$1.00\times$
   & \cellcolor{white}5.34 & \cellcolor{improveGreen}$1.00\times$ \\
  \rowcolor{rowMA}
   & \cellcolor{white}Parallel & \cellcolor{white}16k
   & \cellcolor{white}100.7 & \cellcolor{improveGreen}\textcolor{BrickRed}{$0.86\times$}
   & \cellcolor{white}147.1 & \cellcolor{improveGreen}\textcolor{ForestGreen}{$1.01\times$}
   & \cellcolor{white}3{,}098 & \cellcolor{improveGreen}\textcolor{ForestGreen}{$2.07\times$}
   & \cellcolor{white}11{,}704 & \cellcolor{improveGreen}\textcolor{ForestGreen}{$1.05\times$}
   & \cellcolor{white}5.59 & \cellcolor{improveGreen}\textcolor{BrickRed}{$0.95\times$} \\
  \rowcolor{rowMA}
   & \cellcolor{white}Parallel & \cellcolor{white}8k
   & \cellcolor{white}101.8 & \cellcolor{improveGreen}\textcolor{BrickRed}{$0.85\times$}
   & \cellcolor{white}168.1 & \cellcolor{improveGreen}\textcolor{ForestGreen}{$1.16\times$}
   & \cellcolor{white}5{,}401 & \cellcolor{improveGreen}\textcolor{ForestGreen}{$3.62\times$}
   & \cellcolor{white}11{,}710 & \cellcolor{improveGreen}\textcolor{ForestGreen}{$1.05\times$}
   & \cellcolor{white}3.39 & \cellcolor{improveGreen}\textcolor{ForestGreen}{$1.58\times$} \\
  \rowcolor{rowMA}
   & \cellcolor{white}Parallel & \cellcolor{white}4k
   & \cellcolor{white}105.1 & \cellcolor{improveGreen}\textcolor{BrickRed}{$0.83\times$}
   & \cellcolor{white}205.3 & \cellcolor{improveGreen}\textcolor{ForestGreen}{$1.41\times$}
   & \cellcolor{white}10{,}017 & \cellcolor{improveGreen}\textcolor{ForestGreen}{$6.71\times$}
   & \cellcolor{white}11{,}554 & \cellcolor{improveGreen}\textcolor{ForestGreen}{$1.04\times$}
   & \cellcolor{white}2.12 & \cellcolor{improveGreen}\textcolor{ForestGreen}{$2.52\times$} \\
  \rowcolor{rowMA}
  \multirow{-5}{*}{\textbf{gpt-oss-20B}}
   & \cellcolor{white}Parallel & \cellcolor{white}2k
   & \cellcolor{white}108.3 & \cellcolor{improveGreen}\textcolor{BrickRed}{$0.80\times$}
   & \cellcolor{white}253.8 & \cellcolor{improveGreen}\textcolor{ForestGreen}{$1.75\times$}
   & \cellcolor{white}16{,}180 & \cellcolor{improveGreen}\textcolor{ForestGreen}{$10.84\times$}
   & \cellcolor{white}11{,}303 & \cellcolor{improveGreen}\textcolor{ForestGreen}{$1.02\times$}
   & \cellcolor{white}1.53 & \cellcolor{improveGreen}\textcolor{ForestGreen}{$3.48\times$} \\
  \midrule

  % ─── gpt-oss-120B ──────────────────────────────────────────────────────
  \rowcolor{rowMB}
   & \cellcolor{white}Sequential & \cellcolor{white}—
   & \cellcolor{white}214.5 & \cellcolor{improveGreen}$1.00\times$
   & \cellcolor{white}51.0 & \cellcolor{improveGreen}$1.00\times$
   & \cellcolor{white}1{,}462 & \cellcolor{improveGreen}$1.00\times$
   & \cellcolor{white}9{,}480 & \cellcolor{improveGreen}$1.00\times$
   & \cellcolor{white}7.28 & \cellcolor{improveGreen}$1.00\times$ \\
  \rowcolor{rowMB}
   & \cellcolor{white}Parallel & \cellcolor{white}16k
   & \cellcolor{white}276.0 & \cellcolor{improveGreen}\textcolor{BrickRed}{$0.78\times$}
   & \cellcolor{white}49.8 & \cellcolor{improveGreen}\textcolor{BrickRed}{$0.98\times$}
   & \cellcolor{white}4{,}116 & \cellcolor{improveGreen}\textcolor{ForestGreen}{$2.81\times$}
   & \cellcolor{white}9{,}615 & \cellcolor{improveGreen}\textcolor{ForestGreen}{$1.01\times$}
   & \cellcolor{white}7.64 & \cellcolor{improveGreen}\textcolor{BrickRed}{$0.95\times$} \\
  \rowcolor{rowMB}
   & \cellcolor{white}Parallel & \cellcolor{white}8k
   & \cellcolor{white}270.3 & \cellcolor{improveGreen}\textcolor{BrickRed}{$0.79\times$}
   & \cellcolor{white}59.7 & \cellcolor{improveGreen}\textcolor{ForestGreen}{$1.17\times$}
   & \cellcolor{white}6{,}994 & \cellcolor{improveGreen}\textcolor{ForestGreen}{$4.78\times$}
   & \cellcolor{white}9{,}136 & \cellcolor{improveGreen}\textcolor{BrickRed}{$0.96\times$}
   & \cellcolor{white}5.22 & \cellcolor{improveGreen}\textcolor{ForestGreen}{$1.39\times$} \\
  \rowcolor{rowMB}
   & \cellcolor{white}Parallel & \cellcolor{white}4k
   & \cellcolor{white}217.7 & \cellcolor{improveGreen}\textcolor{BrickRed}{$0.99\times$}
   & \cellcolor{white}90.0 & \cellcolor{improveGreen}\textcolor{ForestGreen}{$1.76\times$}
   & \cellcolor{white}9{,}894 & \cellcolor{improveGreen}\textcolor{ForestGreen}{$6.77\times$}
   & \cellcolor{white}9{,}701 & \cellcolor{improveGreen}\textcolor{ForestGreen}{$1.02\times$}
   & \cellcolor{white}3.58 & \cellcolor{improveGreen}\textcolor{ForestGreen}{$2.04\times$} \\
  \rowcolor{rowMB}
  \multirow{-5}{*}{\textbf{gpt-oss-120B}}
   & \cellcolor{white}Parallel & \cellcolor{white}2k
   & \cellcolor{white}233.1 & \cellcolor{improveGreen}\textcolor{BrickRed}{$0.92\times$}
   & \cellcolor{white}109.0 & \cellcolor{improveGreen}\textcolor{ForestGreen}{$2.14\times$}
   & \cellcolor{white}16{,}486 & \cellcolor{improveGreen}\textcolor{ForestGreen}{$11.28\times$}
   & \cellcolor{white}8{,}913 & \cellcolor{improveGreen}\textcolor{BrickRed}{$0.94\times$}
   & \cellcolor{white}2.51 & \cellcolor{improveGreen}\textcolor{ForestGreen}{$2.91\times$} \\
  \midrule

  % ─── Llama-3.1-8B ──────────────────────────────────────────────────────
  \rowcolor{rowMC}
   & \cellcolor{white}Sequential & \cellcolor{white}—
   & \cellcolor{white}150.8 & \cellcolor{improveGreen}$1.00\times$
   & \cellcolor{white}45.5 & \cellcolor{improveGreen}$1.00\times$
   & \cellcolor{white}1{,}776 & \cellcolor{improveGreen}$1.00\times$
   & \cellcolor{white}5{,}082 & \cellcolor{improveGreen}$1.00\times$
   & \cellcolor{white}14.31 & \cellcolor{improveGreen}$1.00\times$ \\
  \rowcolor{rowMC}
   & \cellcolor{white}Parallel & \cellcolor{white}16k
   & \cellcolor{white}131.7 & \cellcolor{improveGreen}\textcolor{ForestGreen}{$1.14\times$}
   & \cellcolor{white}45.2 & \cellcolor{improveGreen}\textcolor{BrickRed}{$0.99\times$}
   & \cellcolor{white}681 & \cellcolor{improveGreen}\textcolor{BrickRed}{$0.38\times$}
   & \cellcolor{white}5{,}272 & \cellcolor{improveGreen}\textcolor{ForestGreen}{$1.04\times$}
   & \cellcolor{white}32.96 & \cellcolor{improveGreen}\textcolor{BrickRed}{$0.43\times$} \\
  \rowcolor{rowMC}
   & \cellcolor{white}Parallel & \cellcolor{white}8k
   & \cellcolor{white}104.6 & \cellcolor{improveGreen}\textcolor{ForestGreen}{$1.44\times$}
   & \cellcolor{white}62.8 & \cellcolor{improveGreen}\textcolor{ForestGreen}{$1.38\times$}
   & \cellcolor{white}1{,}151 & \cellcolor{improveGreen}\textcolor{BrickRed}{$0.65\times$}
   & \cellcolor{white}5{,}424 & \cellcolor{improveGreen}\textcolor{ForestGreen}{$1.07\times$}
   & \cellcolor{white}17.89 & \cellcolor{improveGreen}\textcolor{BrickRed}{$0.80\times$} \\
  \rowcolor{rowMC}
   & \cellcolor{white}Parallel & \cellcolor{white}4k
   & \cellcolor{white}105.8 & \cellcolor{improveGreen}\textcolor{ForestGreen}{$1.43\times$}
   & \cellcolor{white}71.0 & \cellcolor{improveGreen}\textcolor{ForestGreen}{$1.56\times$}
   & \cellcolor{white}2{,}199 & \cellcolor{improveGreen}\textcolor{ForestGreen}{$1.24\times$}
   & \cellcolor{white}5{,}313 & \cellcolor{improveGreen}\textcolor{ForestGreen}{$1.05\times$}
   & \cellcolor{white}10.43 & \cellcolor{improveGreen}\textcolor{ForestGreen}{$1.37\times$} \\
  \rowcolor{rowMC}
  \multirow{-5}{*}{\textbf{Llama-3.1-8B}}
   & \cellcolor{white}Parallel & \cellcolor{white}2k
   & \cellcolor{white}119.4 & \cellcolor{improveGreen}\textcolor{ForestGreen}{$1.26\times$}
   & \cellcolor{white}83.8 & \cellcolor{improveGreen}\textcolor{ForestGreen}{$1.84\times$}
   & \cellcolor{white}4{,}575 & \cellcolor{improveGreen}\textcolor{ForestGreen}{$2.58\times$}
   & \cellcolor{white}5{,}433 & \cellcolor{improveGreen}\textcolor{ForestGreen}{$1.07\times$}
   & \cellcolor{white}6.57 & \cellcolor{improveGreen}\textcolor{ForestGreen}{$2.18\times$} \\
  \midrule

  % ─── Llama-3.3-70B ─────────────────────────────────────────────────────
  \rowcolor{rowMD}
   & \cellcolor{white}Sequential & \cellcolor{white}—
   & \cellcolor{white}373.5 & \cellcolor{improveGreen}$1.00\times$
   & \cellcolor{white}39.0 & \cellcolor{improveGreen}$1.00\times$
   & \cellcolor{white}8{,}582 & \cellcolor{improveGreen}$1.00\times$
   & \cellcolor{white}5{,}994 & \cellcolor{improveGreen}$1.00\times$
   & \cellcolor{white}22.62 & \cellcolor{improveGreen}$1.00\times$ \\
  \rowcolor{rowMD}
   & \cellcolor{white}Parallel & \cellcolor{white}16k
   & \cellcolor{white}289.7 & \cellcolor{improveGreen}\textcolor{ForestGreen}{$1.29\times$}
   & \cellcolor{white}39.6 & \cellcolor{improveGreen}\textcolor{ForestGreen}{$1.01\times$}
   & \cellcolor{white}5{,}317 & \cellcolor{improveGreen}\textcolor{BrickRed}{$0.62\times$}
   & \cellcolor{white}6{,}145 & \cellcolor{improveGreen}\textcolor{ForestGreen}{$1.03\times$}
   & \cellcolor{white}19.79 & \cellcolor{improveGreen}\textcolor{ForestGreen}{$1.14\times$} \\
  \rowcolor{rowMD}
   & \cellcolor{white}Parallel & \cellcolor{white}8k
   & \cellcolor{white}268.5 & \cellcolor{improveGreen}\textcolor{ForestGreen}{$1.39\times$}
   & \cellcolor{white}46.4 & \cellcolor{improveGreen}\textcolor{ForestGreen}{$1.19\times$}
   & \cellcolor{white}6{,}823 & \cellcolor{improveGreen}\textcolor{BrickRed}{$0.79\times$}
   & \cellcolor{white}5{,}647 & \cellcolor{improveGreen}\textcolor{BrickRed}{$0.94\times$}
   & \cellcolor{white}13.29 & \cellcolor{improveGreen}\textcolor{ForestGreen}{$1.70\times$} \\
  \rowcolor{rowMD}
   & \cellcolor{white}Parallel & \cellcolor{white}4k
   & \cellcolor{white}266.6 & \cellcolor{improveGreen}\textcolor{ForestGreen}{$1.40\times$}
   & \cellcolor{white}52.5 & \cellcolor{improveGreen}\textcolor{ForestGreen}{$1.35\times$}
   & \cellcolor{white}8{,}360 & \cellcolor{improveGreen}\textcolor{BrickRed}{$0.97\times$}
   & \cellcolor{white}5{,}645 & \cellcolor{improveGreen}\textcolor{BrickRed}{$0.94\times$}
   & \cellcolor{white}10.62 & \cellcolor{improveGreen}\textcolor{ForestGreen}{$2.13\times$} \\
  \rowcolor{rowMD}
  \multirow{-5}{*}{\textbf{Llama-3.3-70B}}
   & \cellcolor{white}Parallel & \cellcolor{white}2k
   & \cellcolor{white}292.4 & \cellcolor{improveGreen}\textcolor{ForestGreen}{$1.28\times$}
   & \cellcolor{white}64.1 & \cellcolor{improveGreen}\textcolor{ForestGreen}{$1.64\times$}
   & \cellcolor{white}12{,}303 & \cellcolor{improveGreen}\textcolor{ForestGreen}{$1.43\times$}
   & \cellcolor{white}6{,}427 & \cellcolor{improveGreen}\textcolor{ForestGreen}{$1.07\times$}
   & \cellcolor{white}8.09 & \cellcolor{improveGreen}\textcolor{ForestGreen}{$2.80\times$} \\
  \bottomrule
  \end{tabular}%
  }
\end{table*}

\begin{table*}[!tbp]
  \centering
  \caption{\textbf{LoCoMo: Sequential vs Parallel across block sizes.} Same protocol as Table~\ref{tab:speedup-hotpot}: fixed threshold $\tau=96\text{k}$, Parallel variants sweeping block size from 16k down to 2k tokens. $\Delta$ columns report the speedup over Sequential for the corresponding metric (green improvement, red regression).}
  \label{tab:speedup-locomo}
  \scriptsize
  \setlength{\tabcolsep}{2pt}
  \renewcommand{\arraystretch}{1.15}
  \resizebox{\textwidth}{!}{%
  \begin{tabular}{lll|rc|rc|rc|rc|rc}
  \toprule
  \multirow{2}{*}{\textbf{Model}} & \multirow{2}{*}{\textbf{Variant}} & \multirow{2}{*}{\textbf{Block size}}
   & \multicolumn{2}{c|}{\textbf{E2E Wall (s)} $\downarrow$}
   & \multicolumn{2}{c|}{\textbf{E2E Throughput (tok/s)} $\uparrow$}
   & \multicolumn{2}{c|}{\textbf{Compaction Decode (tok)}}
   & \multicolumn{2}{c|}{\textbf{QA Decode (tok)}}
   & \multicolumn{2}{c}{\textbf{Compaction Throughput (ms/tok)} $\downarrow$} \\
  \cmidrule(lr){4-5}\cmidrule(lr){6-7}\cmidrule(lr){8-9}\cmidrule(lr){10-11}\cmidrule(lr){12-13}
   & & & \cellcolor{white}Value & \cellcolor{captionGreen}\textbf{$\Delta$}
     & \cellcolor{white}Value & \cellcolor{captionGreen}\textbf{$\Delta$}
     & \cellcolor{white}Value & \cellcolor{captionGreen}\textbf{$\Delta$}
     & \cellcolor{white}Value & \cellcolor{captionGreen}\textbf{$\Delta$}
     & \cellcolor{white}Value & \cellcolor{captionGreen}\textbf{$\Delta$} \\
  \midrule
  \rowcolor{sectionBg}\multicolumn{13}{c}{\textit{LoCoMo at$\tau=96$k}} \\
  \midrule

  % ─── gpt-oss-20B ───────────────────────────────────────────────────────
  \rowcolor{rowMA}
   & \cellcolor{white}Sequential & \cellcolor{white}—
   & \cellcolor{white}991.4 & \cellcolor{improveGreen}$1.00\times$
   & \cellcolor{white}134.0 & \cellcolor{improveGreen}$1.00\times$
   & \cellcolor{white}6{,}344 & \cellcolor{improveGreen}$1.00\times$
   & \cellcolor{white}126{,}533 & \cellcolor{improveGreen}$1.00\times$
   & \cellcolor{white}5.23 & \cellcolor{improveGreen}$1.00\times$ \\
  \rowcolor{rowMA}
   & \cellcolor{white}Parallel & \cellcolor{white}16k
   & \cellcolor{white}925.6 & \cellcolor{improveGreen}\textcolor{ForestGreen}{$1.07\times$}
   & \cellcolor{white}138.2 & \cellcolor{improveGreen}\textcolor{ForestGreen}{$1.03\times$}
   & \cellcolor{white}2{,}909 & \cellcolor{improveGreen}\textcolor{BrickRed}{$0.46\times$}
   & \cellcolor{white}125{,}031 & \cellcolor{improveGreen}\textcolor{BrickRed}{$0.99\times$}
   & \cellcolor{white}5.57 & \cellcolor{improveGreen}\textcolor{BrickRed}{$0.94\times$} \\
  \rowcolor{rowMA}
   & \cellcolor{white}Parallel & \cellcolor{white}8k
   & \cellcolor{white}1001.2 & \cellcolor{improveGreen}\textcolor{BrickRed}{$0.99\times$}
   & \cellcolor{white}140.1 & \cellcolor{improveGreen}\textcolor{ForestGreen}{$1.05\times$}
   & \cellcolor{white}5{,}860 & \cellcolor{improveGreen}\textcolor{BrickRed}{$0.92\times$}
   & \cellcolor{white}134{,}431 & \cellcolor{improveGreen}\textcolor{ForestGreen}{$1.06\times$}
   & \cellcolor{white}3.51 & \cellcolor{improveGreen}\textcolor{ForestGreen}{$1.49\times$} \\
  \rowcolor{rowMA}
   & \cellcolor{white}Parallel & \cellcolor{white}4k
   & \cellcolor{white}985.0 & \cellcolor{improveGreen}\textcolor{ForestGreen}{$1.01\times$}
   & \cellcolor{white}145.0 & \cellcolor{improveGreen}\textcolor{ForestGreen}{$1.08\times$}
   & \cellcolor{white}8{,}368 & \cellcolor{improveGreen}\textcolor{ForestGreen}{$1.32\times$}
   & \cellcolor{white}134{,}414 & \cellcolor{improveGreen}\textcolor{ForestGreen}{$1.06\times$}
   & \cellcolor{white}2.42 & \cellcolor{improveGreen}\textcolor{ForestGreen}{$2.16\times$} \\
  \rowcolor{rowMA}
  \multirow{-5}{*}{\textbf{gpt-oss-20B}}
   & \cellcolor{white}Parallel & \cellcolor{white}2k
   & \cellcolor{white}1038.0 & \cellcolor{improveGreen}\textcolor{BrickRed}{$0.96\times$}
   & \cellcolor{white}143.7 & \cellcolor{improveGreen}\textcolor{ForestGreen}{$1.07\times$}
   & \cellcolor{white}16{,}435 & \cellcolor{improveGreen}\textcolor{ForestGreen}{$2.59\times$}
   & \cellcolor{white}132{,}750 & \cellcolor{improveGreen}\textcolor{ForestGreen}{$1.05\times$}
   & \cellcolor{white}1.56 & \cellcolor{improveGreen}\textcolor{ForestGreen}{$3.36\times$} \\
  \midrule

  % ─── gpt-oss-120B ──────────────────────────────────────────────────────
  \rowcolor{rowMB}
   & \cellcolor{white}Sequential & \cellcolor{white}—
   & \cellcolor{white}4905.0 & \cellcolor{improveGreen}$1.00\times$
   & \cellcolor{white}19.9 & \cellcolor{improveGreen}$1.00\times$
   & \cellcolor{white}1{,}479 & \cellcolor{improveGreen}$1.00\times$
   & \cellcolor{white}96{,}196 & \cellcolor{improveGreen}$1.00\times$
   & \cellcolor{white}7.52 & \cellcolor{improveGreen}$1.00\times$ \\
  \rowcolor{rowMB}
   & \cellcolor{white}Parallel & \cellcolor{white}16k
   & \cellcolor{white}3297.0 & \cellcolor{improveGreen}\textcolor{ForestGreen}{$1.49\times$}
   & \cellcolor{white}30.2 & \cellcolor{improveGreen}\textcolor{ForestGreen}{$1.52\times$}
   & \cellcolor{white}1{,}487 & \cellcolor{improveGreen}\textcolor{ForestGreen}{$1.01\times$}
   & \cellcolor{white}98{,}081 & \cellcolor{improveGreen}\textcolor{ForestGreen}{$1.02\times$}
   & \cellcolor{white}19.37 & \cellcolor{improveGreen}\textcolor{BrickRed}{$0.39\times$} \\
  \rowcolor{rowMB}
   & \cellcolor{white}Parallel & \cellcolor{white}8k
   & \cellcolor{white}3065.3 & \cellcolor{improveGreen}\textcolor{ForestGreen}{$1.60\times$}
   & \cellcolor{white}34.4 & \cellcolor{improveGreen}\textcolor{ForestGreen}{$1.73\times$}
   & \cellcolor{white}3{,}769 & \cellcolor{improveGreen}\textcolor{ForestGreen}{$2.55\times$}
   & \cellcolor{white}101{,}795 & \cellcolor{improveGreen}\textcolor{ForestGreen}{$1.06\times$}
   & \cellcolor{white}7.42 & \cellcolor{improveGreen}\textcolor{ForestGreen}{$1.01\times$} \\
  \rowcolor{rowMB}
   & \cellcolor{white}Parallel & \cellcolor{white}4k
   & \cellcolor{white}3601.6 & \cellcolor{improveGreen}\textcolor{ForestGreen}{$1.36\times$}
   & \cellcolor{white}30.7 & \cellcolor{improveGreen}\textcolor{ForestGreen}{$1.54\times$}
   & \cellcolor{white}8{,}308 & \cellcolor{improveGreen}\textcolor{ForestGreen}{$5.62\times$}
   & \cellcolor{white}102{,}150 & \cellcolor{improveGreen}\textcolor{ForestGreen}{$1.06\times$}
   & \cellcolor{white}7.03 & \cellcolor{improveGreen}\textcolor{ForestGreen}{$1.07\times$} \\
  \rowcolor{rowMB}
  \multirow{-5}{*}{\textbf{gpt-oss-120B}}
   & \cellcolor{white}Parallel & \cellcolor{white}2k
   & \cellcolor{white}3851.9 & \cellcolor{improveGreen}\textcolor{ForestGreen}{$1.27\times$}
   & \cellcolor{white}32.1 & \cellcolor{improveGreen}\textcolor{ForestGreen}{$1.61\times$}
   & \cellcolor{white}17{,}816 & \cellcolor{improveGreen}\textcolor{ForestGreen}{$12.05\times$}
   & \cellcolor{white}105{,}808 & \cellcolor{improveGreen}\textcolor{ForestGreen}{$1.10\times$}
   & \cellcolor{white}4.49 & \cellcolor{improveGreen}\textcolor{ForestGreen}{$1.68\times$} \\
  \midrule

  % ─── Llama-3.1-8B ──────────────────────────────────────────────────────
  \rowcolor{rowMC}
   & \cellcolor{white}Sequential & \cellcolor{white}—
   & \cellcolor{white}495.2 & \cellcolor{improveGreen}$1.00\times$
   & \cellcolor{white}52.4 & \cellcolor{improveGreen}$1.00\times$
   & \cellcolor{white}542 & \cellcolor{improveGreen}$1.00\times$
   & \cellcolor{white}25{,}406 & \cellcolor{improveGreen}$1.00\times$
   & \cellcolor{white}12.80 & \cellcolor{improveGreen}$1.00\times$ \\
  \rowcolor{rowMC}
   & \cellcolor{white}Parallel & \cellcolor{white}16k
   & \cellcolor{white}555.9 & \cellcolor{improveGreen}\textcolor{BrickRed}{$0.89\times$}
   & \cellcolor{white}55.5 & \cellcolor{improveGreen}\textcolor{ForestGreen}{$1.06\times$}
   & \cellcolor{white}616 & \cellcolor{improveGreen}\textcolor{ForestGreen}{$1.14\times$}
   & \cellcolor{white}30{,}212 & \cellcolor{improveGreen}\textcolor{ForestGreen}{$1.19\times$}
   & \cellcolor{white}32.82 & \cellcolor{improveGreen}\textcolor{BrickRed}{$0.39\times$} \\
  \rowcolor{rowMC}
   & \cellcolor{white}Parallel & \cellcolor{white}8k
   & \cellcolor{white}530.1 & \cellcolor{improveGreen}\textcolor{BrickRed}{$0.93\times$}
   & \cellcolor{white}53.2 & \cellcolor{improveGreen}\textcolor{ForestGreen}{$1.02\times$}
   & \cellcolor{white}1{,}198 & \cellcolor{improveGreen}\textcolor{ForestGreen}{$2.21\times$}
   & \cellcolor{white}27{,}010 & \cellcolor{improveGreen}\textcolor{ForestGreen}{$1.06\times$}
   & \cellcolor{white}18.28 & \cellcolor{improveGreen}\textcolor{BrickRed}{$0.70\times$} \\
  \rowcolor{rowMC}
   & \cellcolor{white}Parallel & \cellcolor{white}4k
   & \cellcolor{white}594.6 & \cellcolor{improveGreen}\textcolor{BrickRed}{$0.83\times$}
   & \cellcolor{white}60.1 & \cellcolor{improveGreen}\textcolor{ForestGreen}{$1.15\times$}
   & \cellcolor{white}2{,}245 & \cellcolor{improveGreen}\textcolor{ForestGreen}{$4.14\times$}
   & \cellcolor{white}33{,}469 & \cellcolor{improveGreen}\textcolor{ForestGreen}{$1.32\times$}
   & \cellcolor{white}11.06 & \cellcolor{improveGreen}\textcolor{ForestGreen}{$1.16\times$} \\
  \rowcolor{rowMC}
  \multirow{-5}{*}{\textbf{Llama-3.1-8B}}
   & \cellcolor{white}Parallel & \cellcolor{white}2k
   & \cellcolor{white}607.0 & \cellcolor{improveGreen}\textcolor{BrickRed}{$0.82\times$}
   & \cellcolor{white}63.4 & \cellcolor{improveGreen}\textcolor{ForestGreen}{$1.21\times$}
   & \cellcolor{white}4{,}357 & \cellcolor{improveGreen}\textcolor{ForestGreen}{$8.04\times$}
   & \cellcolor{white}34{,}110 & \cellcolor{improveGreen}\textcolor{ForestGreen}{$1.34\times$}
   & \cellcolor{white}6.81 & \cellcolor{improveGreen}\textcolor{ForestGreen}{$1.88\times$} \\
  \midrule

  % ─── Llama-3.3-70B ─────────────────────────────────────────────────────
  \rowcolor{rowMD}
   & \cellcolor{white}Sequential & \cellcolor{white}—
   & \cellcolor{white}1194.5 & \cellcolor{improveGreen}$1.00\times$
   & \cellcolor{white}38.9 & \cellcolor{improveGreen}$1.00\times$
   & \cellcolor{white}1{,}451 & \cellcolor{improveGreen}$1.00\times$
   & \cellcolor{white}45{,}043 & \cellcolor{improveGreen}$1.00\times$
   & \cellcolor{white}22.32 & \cellcolor{improveGreen}$1.00\times$ \\
  \rowcolor{rowMD}
   & \cellcolor{white}Parallel & \cellcolor{white}16k
   & \cellcolor{white}1495.6 & \cellcolor{improveGreen}\textcolor{BrickRed}{$0.80\times$}
   & \cellcolor{white}41.5 & \cellcolor{improveGreen}\textcolor{ForestGreen}{$1.07\times$}
   & \cellcolor{white}6{,}708 & \cellcolor{improveGreen}\textcolor{ForestGreen}{$4.62\times$}
   & \cellcolor{white}55{,}359 & \cellcolor{improveGreen}\textcolor{ForestGreen}{$1.23\times$}
   & \cellcolor{white}22.12 & \cellcolor{improveGreen}\textcolor{ForestGreen}{$1.01\times$} \\
  \rowcolor{rowMD}
   & \cellcolor{white}Parallel & \cellcolor{white}8k
   & \cellcolor{white}1306.8 & \cellcolor{improveGreen}\textcolor{BrickRed}{$0.91\times$}
   & \cellcolor{white}40.0 & \cellcolor{improveGreen}\textcolor{ForestGreen}{$1.03\times$}
   & \cellcolor{white}4{,}179 & \cellcolor{improveGreen}\textcolor{ForestGreen}{$2.88\times$}
   & \cellcolor{white}48{,}050 & \cellcolor{improveGreen}\textcolor{ForestGreen}{$1.07\times$}
   & \cellcolor{white}17.58 & \cellcolor{improveGreen}\textcolor{ForestGreen}{$1.27\times$} \\
  \rowcolor{rowMD}
   & \cellcolor{white}Parallel & \cellcolor{white}4k
   & \cellcolor{white}1275.2 & \cellcolor{improveGreen}\textcolor{BrickRed}{$0.94\times$}
   & \cellcolor{white}42.1 & \cellcolor{improveGreen}\textcolor{ForestGreen}{$1.08\times$}
   & \cellcolor{white}8{,}494 & \cellcolor{improveGreen}\textcolor{ForestGreen}{$5.85\times$}
   & \cellcolor{white}45{,}193 & \cellcolor{improveGreen}\textcolor{ForestGreen}{$1.00\times$}
   & \cellcolor{white}10.27 & \cellcolor{improveGreen}\textcolor{ForestGreen}{$2.17\times$} \\
  \rowcolor{rowMD}
  \multirow{-5}{*}{\textbf{Llama-3.3-70B}}
   & \cellcolor{white}Parallel & \cellcolor{white}2k
   & \cellcolor{white}1388.7 & \cellcolor{improveGreen}\textcolor{BrickRed}{$0.86\times$}
   & \cellcolor{white}50.9 & \cellcolor{improveGreen}\textcolor{ForestGreen}{$1.31\times$}
   & \cellcolor{white}23{,}009 & \cellcolor{improveGreen}\textcolor{ForestGreen}{$15.86\times$}
   & \cellcolor{white}47{,}655 & \cellcolor{improveGreen}\textcolor{ForestGreen}{$1.06\times$}
   & \cellcolor{white}6.81 & \cellcolor{improveGreen}\textcolor{ForestGreen}{$3.28\times$} \\
  \bottomrule
  \end{tabular}%
  }
\end{table*}

\FloatBarrier

Table~\ref{tab:matched-decode-tpot} therefore selects runs from the full sweep where Sequential and Parallel produce comparable compaction decode volumes, isolating the throughput improvement attributable to parallelism rather than token count differences.

% Requires in preamble:
%   \usepackage[dvipsnames,table]{xcolor}
%   \usepackage{multirow}
%   \usepackage{booktabs}
% Colors expected from main.tex: improveGreen, captionGreen, rowMA..rowMD.

\begin{table}[H]
    \centering
%    \captionsetup{font=footnotesize}
    \caption{Parallel achieves lower TPOT at matched decode volume.}
    \label{tab:matched-decode-tpot}
    \scriptsize
    \setlength{\tabcolsep}{4pt}
    \renewcommand{\arraystretch}{1.2}
    \resizebox{\columnwidth}{!}{%
    \begin{tabular}{c l l c r r c}
    \toprule
    \textbf{\#} & \textbf{Model} & \textbf{Benchmark} & \textbf{Block}
     & \textbf{Seq.\ (tok)} & \textbf{Par.\ (tok)}
     & \cellcolor{captionGreen}\textbf{$\Delta$ Throughput} \\
    \midrule
    \rowcolor{rowMD}
    1 & Llama-3.3-70B & HotpotQA & 4\,k & 8{,}582 & 8{,}360 & \cellcolor{improveGreen}\textcolor{ForestGreen}{$2.13\times$} \\
    \rowcolor{rowMD}
    2 & Llama-3.3-70B & HotpotQA & 8\,k & 8{,}582 & 6{,}823 & \cellcolor{improveGreen}\textcolor{ForestGreen}{$1.70\times$} \\
    \rowcolor{rowMA}
    3 & gpt-oss-20B   & LoCoMo   & 8\,k & 6{,}344 & 5{,}860 & \cellcolor{improveGreen}\textcolor{ForestGreen}{$1.49\times$} \\
    \rowcolor{rowMC}
    4 & Llama-3.1-8B  & HotpotQA & 4\,k & 1{,}776 & 2{,}199 & \cellcolor{improveGreen}\textcolor{ForestGreen}{$1.37\times$} \\
    \bottomrule
    \end{tabular}%
    }
\end{table}

\noindent \textbf{Prefill cost limits parallelism.} Average compaction throughput measures the combined prefill and decode latency per output token during compaction. In the parallel setting, Workers share the common conversation prefix via the prefix cache, but each worker must additionally prefill its own unique block content, which is not cached. At the 16\,k block size, this uncached portion is 16\,k tokens per worker, incurring substantial prefill overhead that the parallelism gain cannot compensate for, resulting in consistently worse throughput compared to smaller block sizes.

\begin{figure}[h]
\centering
\includegraphics[width=0.9\columnwidth]{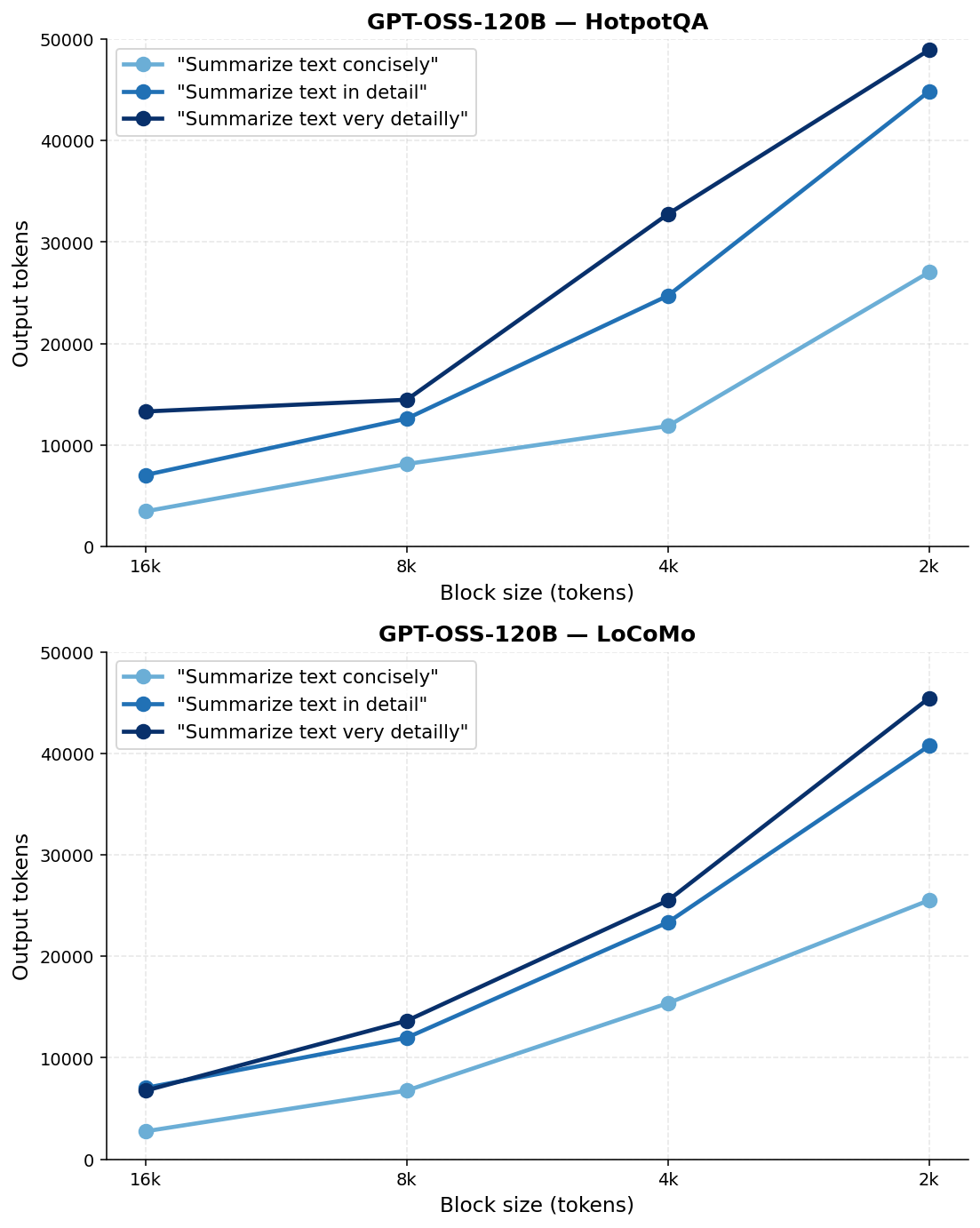}
%\captionsetup{font=footnotesize}
\caption{GPT-OSS-120B output tokens vs.\ block size across three prompt variants on HotpotQA (left) and LoCoMo (right).
}\vspace{-4mm}
\label{fig:gptoss120-output-tokens}
\end{figure}

\noindent \textbf{Prompt sensitivity under parallel compaction.} While sequential compaction largely ignores prompt length instructions, parallel compaction restores some prompt sensitivity. Figure~\ref{fig:gptoss120-output-tokens} shows gpt-oss-120B output tokens across three prompt variants and all block sizes on HotpotQA and LoCoMo; other backbones exhibit similar trends. The model does respond to the prompt within each block size, with more detailed prompts producing more tokens. Combined with block size, this gives the operator two independent levers: block size controls the coarse volume of compaction output, while the prompt can further tune output length within each block. 

\begin{table}[h]
    \centering
    %\captionsetup{font=footnotesize}
    \caption{gpt-oss-120B compaction output / 96\,k input ratio (\%).}
    \label{tab:gptoss120-out-in-ratio}
    \scriptsize
    \setlength{\tabcolsep}{2pt}
    \renewcommand{\arraystretch}{1.15}
    \resizebox{\columnwidth}{!}{%
    \begin{tabular}{l|ccc|ccc}
    \toprule
    \multirow{2}{*}{\textbf{Config}}
      & \multicolumn{3}{c|}{\textbf{HotpotQA}}
      & \multicolumn{3}{c}{\textbf{LoCoMo}} \\
    \cmidrule(lr){2-4}\cmidrule(lr){5-7}
     & Concise & Detailly & Very Detailly
     & Concise & Detailly & Very Detailly \\
    \midrule
    Sequential & 0.79\%  & 2.98\%  & 4.16\%  & 0.92\%  & 3.74\%  & 4.05\% \\
    16k        & 3.62\%  & 7.35\%  & 13.87\% & 2.87\%  & 7.32\%  & 7.05\% \\
    8k         & 8.48\%  & 13.14\% & 15.07\% & 7.05\%  & 12.47\% & 14.22\% \\
    4k         & 12.37\% & 25.75\% & 34.13\% & 16.02\% & 24.32\% & 26.57\% \\
    2k         & 28.16\% & 46.73\% & 50.98\% & 26.59\% & 42.44\% & 47.34\% \\
    \bottomrule
    \end{tabular}%
    }
\end{table}

As shown in Table~\ref{tab:gptoss120-out-in-ratio}, this two-axis control is quantitatively meaningful. By tuning the block size alone, the operator can scale compaction output to roughly 25\% of the original context, and by additionally selecting a more detailed prompt, the output can reach up to 50\%, offering fine-grained, predictable control over how much context survives each compaction call. 

\noindent \textbf{Output token tradeoff.} While the results above demonstrate that output volume can scale up to 50\% of the original context, retaining that much is not the goal in itself. The operator can equally scale down by choosing a concise prompt, reducing output significantly. The key insight is bidirectional control: block size and prompt together allow the operator to tune compaction volume precisely in either direction. Keeping more tokens in the summary is not always better: longer summaries re-introduce the same attention cost and context rot effects that motivated compaction in the first place. As the retained context grows, the model must again attend over a longer, lower-signal history, degrading reasoning quality before the next compaction trigger. The optimal operating point is therefore not the maximum but the minimum sufficient to preserve task-relevant information, and the two-axis control allows the operator to target that point precisely.  

\takeaway{Whether the technique is compaction, token-efficient tool design, or just-in-time retrieval, recent practitioner guidance converges on a common principle: keep inside the context window only the smallest set of high-signal tokens likely to maximize the desired outcome~\cite{anthropic2025effectivecontext}.}

\section{Related Work}
\noindent  \textbf{Context management.} Managing the growing context of long-horizon LLM agents has been studied along several axes~\cite{liu2024lost, kang2025acon}. Existing approaches range from simple truncation to summarization-based compaction and complex retrieval-based memory systems. It has been shown that truncation-based approaches are insufficient for maintaining coherence in long conversations, as summarization methods significantly outperform sliding window baselines~\cite{xu2022beyond}. Token-level compression methods reduce context length by removing low-information tokens, but these approaches lack semantic awareness and disrupt coherence in multi-step reasoning chains~\cite{jiang2023llmlingua, jiang2024longllmlingua}. Retrieval-based approaches such as MemGPT store conversation history in external memory and retrieve relevant segments on demand, but require dedicated retrieval infrastructure and add latency for each lookup~\cite{packer2023memgpt}.

\noindent \textbf{Latent space compression.} KV-cache and latent-space compression methods offer complementary approaches: Cartridges~\cite{eyuboglu2025cartridges} pre-train lightweight KV-cache representations offline to reduce inference cost on large corpora, while Zweiger et al.~\cite{zweiger2026fast} compress KV caches at runtime via attention-pattern matching. However, Cartridges require offline training and are not applicable at runtime, while KV-cache compression operates in the latent space, making the compressed representation opaque to humans, non-transferable across models, and dependent on white-box access to the model's internal cache. LLM-based summarization, by contrast, produces human-readable text that is model-agnostic and requires no access to model internals.

\noindent \textbf{Effective context length.} Beyond keeping the conversation within the model's hard context limit, recent work suggests that keeping fewer, higher-signal tokens inside the window yields better downstream accuracy, motivating more frequent compression of the active context. One study~\cite{modarressi2025nolima} shows that across 13 models claiming 128k to 2M token windows, the effective context length (the length at which accuracy stays within 85\% of the short-context baseline) is only between roughly 1k and 8k tokens. Another study~\cite{du2025context} sharpens this by controlling for retrieval: performance still degrades substantially with input length even when the model retrieves all evidence with exact match, and even when non-evidence tokens are attention-masked, establishing input length itself as a first-order cause of reasoning degradation. These results reframe compaction not just as a mechanism to fit within the context window, but as a primary lever for keeping the active context inside the regime where attention functions effectively.

\noindent \textbf{Parallel document summarization.} Divide-and-conquer approaches for long document summarization split the input into chunks, summarize each independently, and merge the results~\cite{zhou2024llm, zhang2024chain, ou2025context,xiao2025long}. Recent theoretical work characterizes when such chunked processing outperforms single-pass approaches~\cite{xu2025does}. Our work differs in targeting runtime context compaction in agentic flows, parallelizing summarization via prefix-aware blocks where each block sees the full conversation history preceding it, preserving cross-block context dependencies, with only a small additional cost from placing a target marker at the end of each prefix, while enabling prefix cache reuse across workers.

\noindent \textbf{LLMs ignore length instructions.} Multiple benchmarks have shown that LLMs systematically fail to follow explicit length constraints: models struggle to match specified word or token counts, especially at longer targets~\cite{zhang2026lifebench,akinfaderin2025plan,xie2025prompt}, and performance degrades further in long-context scenarios~\cite{wu2025lifbench}. LLMs also exhibit verbosity compensation, generating excessively long outputs when uncertain~\cite{zhang2025demystify}, and lack the ability to track their own output length during generation~\cite{xiao2026can}. Evaluation of long-form generation confirms these limitations persist even when models are explicitly instructed~\cite{wu2025longgenbench}. These findings support our design choice: since prompt instructions cannot reliably control summary volume, we use the number of parallel workers as the primary control knob for compaction output. Unlike prior work, we target runtime agentic conversation compaction with no offline training, retrieval infrastructure, or white-box model access, and introduce a prefix-aware target-at-end design that preserves cross-block context and enables prefix cache reuse.

\section{Discussion and Future Work}
\label{sec:discussion}

\noindent \textbf{Dynamic block size and prompt selection.} In this work, we use a fixed block size and a single compaction prompt. A natural extension is to adapt both per block based on content type: tool call outputs with shallow structure could use larger blocks and a concise prompt, while math-heavy or insight-dense segments could use smaller blocks with a detailed prompt, steering compaction granularity and verbosity where it matters most.

\noindent \textbf{Model fine-tuning for better marker awareness.} Our prefix-aware layout inserts XML markers to delimit block boundaries. Current models were not trained to attend to these markers during summarization. Fine-tuning on marker-annotated compaction traces could improve boundary awareness and produce more faithful per-block summaries.

\noindent \textbf{Asynchronous compaction.} The current design blocks the agent until all workers finish. An async variant would run block workers in the background while the main thread continues serving inference on the residual context, merging the completed summaries once the current turn finishes. Tool calls offer a natural trigger: whenever the agent is waiting on \texttt{pytest}, a shell command, or any external API, the GPU is idle, and compaction can proceed at zero opportunity cost.  

\section{Conclusion} 
In this work, we characterized sequential and parallel context compaction for long-horizon LLM agents across four backbones and two benchmarks. Our detailed characterization reveals that summarization output is largely input- and prompt-invariant, making prompt engineering an unreliable control knob for summary volume. We introduced parallel block compaction, which gives the operator direct control over summary volume through block count and increases compaction throughput even at matched decode volume.

\FloatBarrier
\newpage
\bibliographystyle{IEEEtranS}
\bibliography{reference}

@misc{anthropic2025claudecode,
  title={Claude Code: Best practices for agentic coding},
  author={{Anthropic}},
  year={2025},
  url={https://platform.claude.com/docs/en/build-with-claude/compaction}
}

@misc{openai2025codex,
  title        = {Codex Prompting Guide},
  author       = {{OpenAI}},
  year         = {2025},
  note         = {[Online]. Available: \url{https://developers.openai.com/cookbook/examples/gpt-5/codex_prompting_guide}}
}

@misc{langchain,
  title={LangChain: Building applications with LLMs through composability},
  author={{LangChain}},
  year={2023},
  url={https://github.com/langchain-ai/langchain}
}

@misc{llamaindex,
  title={LlamaIndex: A data framework for LLM applications},
  author={{LlamaIndex}},
  year={2023},
  url={https://github.com/run-llama/llama_index}
}

@article{liu2024lost,
  title={Lost in the middle: How language models use long contexts},
  author={Liu, Nelson F and Lin, Kevin and Hewitt, John and Paranjape, Ashwin and Bevilacqua, Michele and Petroni, Fabio and Liang, Percy},
  journal={Transactions of the association for computational linguistics},
  volume={12},
  pages={157--173},
  year={2024}
}

@techreport{hong2025contextrot,
  title = {Context Rot: How Increasing Input Tokens Impacts LLM Performance},
  author = {Hong, Kelly and Troynikov, Anton and Huber, Jeff},
  institution = {Chroma},
  year = {2025},
  month = {July},
  url = {https://research.trychroma.com/context-rot}
}

@article{kang2025acon,
  title={Acon: Optimizing context compression for long-horizon llm agents},
  author={Kang, Minki and Chen, Wei-Ning and Han, Dongge and Inan, Huseyin A and Wutschitz, Lukas and Chen, Yanzhi and Sim, Robert and Rajmohan, Saravan},
  journal={arXiv preprint arXiv:2510.00615},
  year={2025}
}

@inproceedings{xu2022beyond,
  title={Beyond goldfish memory: Long-term open-domain conversation},
  author={Xu, Jing and Szlam, Arthur and Weston, Jason},
  booktitle={Proceedings of the 60th annual meeting of the association for computational linguistics (volume 1: long papers)},
  pages={5180--5197},
  year={2022}
}

@article{packer2023memgpt,
  title={MemGPT: towards LLMs as operating systems.},
  author={Packer, Charles and Fang, Vivian and Patil, Shishir\_G and Lin, Kevin and Wooders, Sarah and Gonzalez, Joseph\_E},
  year={2023},
  publisher={ArXiv}
}

@inproceedings{jiang2024longllmlingua,
  title={Longllmlingua: Accelerating and enhancing llms in long context scenarios via prompt compression},
  author={Jiang, Huiqiang and Wu, Qianhui and Luo, Xufang and Li, Dongsheng and Lin, Chin-Yew and Yang, Yuqing and Qiu, Lili},
  booktitle={Proceedings of the 62nd Annual Meeting of the Association for Computational Linguistics (Volume 1: Long Papers)},
  pages={1658--1677},
  year={2024}
}

@inproceedings{jiang2023llmlingua,
  title={Llmlingua: Compressing prompts for accelerated inference of large language models},
  author={Jiang, Huiqiang and Wu, Qianhui and Lin, Chin-Yew and Yang, Yuqing and Qiu, Lili},
  booktitle={Proceedings of the 2023 conference on empirical methods in natural language processing},
  pages={13358--13376},
  year={2023}
}

@article{eyuboglu2025cartridges,
  title={Cartridges: Lightweight and general-purpose long context representations via self-study},
  author={Eyuboglu, Sabri and Ehrlich, Ryan and Arora, Simran and Guha, Neel and Zinsley, Dylan and Liu, Emily and Tennien, Will and Rudra, Atri and Zou, James and Mirhoseini, Azalia and others},
  journal={arXiv preprint arXiv:2506.06266},
  year={2025}
}

@article{zweiger2026fast,
  title={Fast kv compaction via attention matching},
  author={Zweiger, Adam and Fu, Xinghong and Guo, Han and Kim, Yoon},
  journal={arXiv preprint arXiv:2602.16284},
  year={2026}
}

@article{modarressi2025nolima,
  title={Nolima: Long-context evaluation beyond literal matching},
  author={Modarressi, Ali and Deilamsalehy, Hanieh and Dernoncourt, Franck and Bui, Trung and Rossi, Ryan A and Yoon, Seunghyun and Sch{\"u}tze, Hinrich},
  journal={arXiv preprint arXiv:2502.05167},
  year={2025}
}

@article{du2025context,
  title={Context length alone hurts LLM performance despite perfect retrieval},
  author={Du, Yufeng and Tian, Minyang and Ronanki, Srikanth and Rongali, Subendhu and Bodapati, Sravan and Galstyan, Aram and Wells, Azton and Schwartz, Roy and Huerta, Eliu A and Peng, Hao},
  journal={arXiv preprint arXiv:2510.05381},
  year={2025}
}

@inproceedings{ou2025context,
  title={Context-aware hierarchical merging for long document summarization},
  author={Ou, Litu and Lapata, Mirella},
  booktitle={Findings of the Association for Computational Linguistics: ACL 2025},
  pages={5534--5561},
  year={2025}
}

@misc{zhou2024llm,
title={LLM$\times$MapReduce: Simplified Long-Sequence Processing using Large Language Models}, 
author={Zihan Zhou and Chong Li and Xinyi Chen and Shuo Wang and Yu Chao and Zhili Li and Haoyu Wang and Rongqiao An and Qi Shi and Zhixing Tan and Xu Han and Xiaodong Shi and Zhiyuan Liu and Maosong Sun},
year={2024},
eprint={2410.09342},
archivePrefix={arXiv},
primaryClass={cs.CL},
url={https://arxiv.org/abs/2410.09342}, 
}

@article{zhang2024chain,
  title={Chain of agents: Large language models collaborating on long-context tasks},
  author={Zhang, Yusen and Sun, Ruoxi and Chen, Yanfei and Pfister, Tomas and Zhang, Rui and Ar{\i}k, Sercan {\"O}},
  journal={Advances in Neural Information Processing Systems},
  volume={37},
  pages={132208--132237},
  year={2024}
}

@article{xu2025does,
  title={When Does Divide and Conquer Work for Long Context LLM? A Noise Decomposition Framework},
  author={Xu, Zhen and Zhu, Shang and Wang, Jue and Wang, Junlin and Athiwaratkun, Ben and Wang, Chi and Zou, James and Zhang, Ce},
  journal={arXiv preprint arXiv:2506.16411},
  year={2025}
}

@inproceedings{zhang2025demystify,
  title={Demystify Verbosity Compensation Behavior of Large Language Models},
  author={Zhang, Yusen and Das, Sarkar Snigdha Sarathi and Zhang, Rui},
  booktitle={Proceedings of the 2nd Workshop on Uncertainty-Aware NLP (UncertaiNLP 2025)},
  pages={160--178},
  year={2025}
}

@article{xiao2026can,
  title={Can LLMs Track Their Output Length? A Dynamic Feedback Mechanism for Precise Length Regulation},
  author={Xiao, Meiman and Wang, Ante and Hu, Qingguo and Miao, Zhongjian and Shen, Huangjun and Wang, Longyue and Luo, Weihua and Su, Jinsong},
  journal={arXiv preprint arXiv:2601.01768},
  year={2026}
}

@inproceedings{wu2025longgenbench,
  title={Longgenbench: Benchmarking long-form generation in long context llms},
  author={Wu, Yuhao and Hee, Ming Shan and Hu, Zhiqiang and Lee, Roy Ka-Wei},
  booktitle={International Conference on Learning Representations},
  volume={2025},
  pages={6851--6872},
  year={2025}
}

@article{zhang2026lifebench,
  title={Lifebench: Evaluating length instruction following in large language models},
  author={Zhang, Wei and Zhou, Zhenhong and Wang, Kun and Fang, Junfeng and Xu, Rongwu and Zhang, Yuanhe and Wang, Rui and Zhang, Ge and Li, Xinfeng and Sun, Li and others},
  journal={Advances in Neural Information Processing Systems},
  volume={38},
  year={2026}
}

@inproceedings{wu2025lifbench,
  title={Lifbench: Evaluating the instruction following performance and stability of large language models in long-context scenarios},
  author={Wu, Xiaodong and Wang, Minhao and Liu, Yichen and Shi, Xiaoming and Yan, He and Xiangju, Lu and Zhu, Junmin and Zhang, Wei},
  booktitle={Proceedings of the 63rd Annual Meeting of the Association for Computational Linguistics (Volume 1: Long Papers)},
  pages={16445--16468},
  year={2025}
}

@inproceedings{yang2018hotpotqa,
  title={HotpotQA: A dataset for diverse, explainable multi-hop question answering},
  author={Yang, Zhilin and Qi, Peng and Zhang, Saizheng and Bengio, Yoshua and Cohen, William and Salakhutdinov, Ruslan and Manning, Christopher D},
  booktitle={Proceedings of the 2018 conference on empirical methods in natural language processing},
  pages={2369--2380},
  year={2018}
}

@inproceedings{maharana2024evaluating,
  title={Evaluating very long-term conversational memory of llm agents},
  author={Maharana, Adyasha and Lee, Dong-Ho and Tulyakov, Sergey and Bansal, Mohit and Barbieri, Francesco and Fang, Yuwei},
  booktitle={Proceedings of the 62nd Annual Meeting of the Association for Computational Linguistics (Volume 1: Long Papers)},
  pages={13851--13870},
  year={2024}
}

@article{panickssery2024llm,
  title={Llm evaluators recognize and favor their own generations},
  author={Panickssery, Arjun and Bowman, Samuel R and Feng, Shi},
  journal={Advances in Neural Information Processing Systems},
  volume={37},
  pages={68772--68802},
  year={2024}
}

@inproceedings{xu2024pride,
  title={Pride and prejudice: LLM amplifies self-bias in self-refinement},
  author={Xu, Wenda and Zhu, Guanglei and Zhao, Xuandong and Pan, Liangming and Li, Lei and Wang, William},
  booktitle={Proceedings of the 62nd Annual Meeting of the Association for Computational Linguistics (Volume 1: Long Papers)},
  pages={15474--15492},
  year={2024}
}

@misc{percena_locomo_mc10,
  title        = {{LoCoMo-MC10}: A 10-choice Multiple-Choice Version of LoCoMo},
  author       = {{Percena}},
  year         = {2026},
  howpublished = {\url{https://huggingface.co/datasets/Percena/locomo-mc10}},
  note         = {Hugging Face dataset. Accessed: 2026-05-19}
}

@misc{anthropic2025effectivecontext,
  title        = {Effective Context Engineering for AI Agents},
  author       = {{Anthropic}},
  year         = {2025},
  howpublished = {\url{https://www.anthropic.com/engineering/effective-context-engineering-for-ai-agents}},
  note         = {Accessed: 2026-05-11}
}

@article{xiao2025long,
  title={Long context scaling: Divide and conquer via multi-agent question-driven collaboration},
  author={Xiao, Sibo and Lin, Zixin and Gao, Wenyang and Chen, Hui and Zhang, Yue},
  journal={arXiv preprint arXiv:2505.20625},
  year={2025}
}

@inproceedings{ravaut2024context,
  title={On context utilization in summarization with large language models},
  author={Ravaut, Mathieu and Sun, Aixin and Chen, Nancy and Joty, Shafiq},
  booktitle={Proceedings of the 62nd Annual Meeting of the Association for Computational Linguistics (Volume 1: Long Papers)},
  pages={2764--2781},
  year={2024}
}

@article{akinfaderin2025plan,
  title={Plan-and-Write: Structure-Guided Length Control for LLMs without Model Retraining},
  author={Akinfaderin, Adewale and Subramanian, Shreyas and Sehwag, Akarsha},
  journal={arXiv preprint arXiv:2511.01807},
  year={2025}
}

@article{xie2025prompt,
  title={Prompt-Based One-Shot Exact Length-Controlled Generation with LLMs},
  author={Xie, Juncheng and Lee, Hung-yi},
  journal={arXiv preprint arXiv:2508.13805},
  year={2025}
}

\end{document}